\documentclass[journal]{IEEEtran}
\usepackage[utf8]{inputenc}
\usepackage{amsmath,amssymb,amsfonts}
\usepackage[nospace]{cite}

\usepackage{algorithmic}
\usepackage{multirow}
\usepackage{graphicx}
\usepackage{textcomp}
\usepackage{xcolor}
\usepackage{pifont}
\usepackage{amsthm}
\usepackage{hyperref}
\usepackage{mathrsfs}
\usepackage{booktabs}
\usepackage{hyperref}
\usepackage{cases}
\usepackage{comment}
\usepackage{empheq}
\usepackage{caption}
\usepackage{subcaption}
\usepackage{fancyhdr}
\usepackage{tabularx}
\usepackage{tikz}
\usepackage{multirow}

\allowdisplaybreaks

\usepackage{url}  %Required
\usepackage{graphicx}  %Required
\usepackage[ruled,vlined,linesnumbered]{algorithm2e}
\usepackage{setspace}
\usepackage{floatpag}
\usepackage{float}
\floatpagestyle{empty}

\usepackage{listings}

\begin{document}

\title{Sparse Attention-driven Quality Prediction for Production Process Optimization in Digital Twins}

\author{
\IEEEauthorblockN{Yanlei Yin, 
Lihua Wang, 
Dinh Thai Hoang,~\IEEEmembership{Senior Member,~IEEE}, 
Wenbo~Wang,~\IEEEmembership{Senior Member,~IEEE} 
and Dusit Niyato,~\IEEEmembership{Fellow,~IEEE}
}

\thanks{This research is supported in part by National Natural Science Foundation of China under grant No. 62302046, in part by National Natural Science Foundation of China under Yunnan Major Scientific and Technological Projects under Grant 202202AG050002, in part by the National Key Research and Development Program of China under grant No. 2023YFB3308401, in part by ChingMu Tech. Ltd. Research Project ``WiTracker’’ under grant No. KKF0202301252, in part 
by the National Research Foundation, Singapore, and Infocomm Media Development Authority under its Future Communications Research \& Development Programme, in part by Defence Science Organisation (DSO) National Laboratories under the AI Singapore Programme (FCP-NTU-RG-2022-010 and FCP-ASTAR-TG-2022-003), in part by Singapore Ministry of Education (MOE) Tier 1 (RG87/22), and in part by the NTU Centre for Computational Technologies in Finance (NTU-CCTF).
}

\thanks{Yanlei Yin and Lihua Wang are with the School of Mechanical and Electrical Engineering, Kunming University of Science and Technology, Kunming, China, 650500 (e-mails: \mbox{20201103005@stu.kust.edu.cn}, \mbox{wanglihua@kust.edu.cn}).}

\thanks{Dinh Thai Hoang is with the School of Electrical and Data Engineering, University of Technology Sydney (UTS), Australia (email: \mbox{Hoang.Dinh@uts.edu.au}). }

\thanks{Wenbo Wang is with the School of Mechanical and Electrical Engineering, and Yunnan Key Laboratory of Intelligent Control and Application, Kunming University of Science and Technology, Kunming, China, 650500 (e-mail: \mbox{wenbo\_wang@kust.edu.cn}). (corresponding author)}

\thanks{Dusit Niyato is with the College of Computing and Data Science, Nanyang Technological University, Singapore, 639798 (email: \mbox{dniyato@ntu.edu.sg}).}
}

\maketitle

\begin{abstract}
In the process industry, long-term and efficient optimization of production lines requires real-time monitoring and analysis of operational states to fine-tune production line parameters. However, complexity in operational logic and intricate coupling of production process parameters make it difficult to develop an accurate mathematical model for the entire process, thus hindering the deployment of efficient optimization mechanisms. In view of these difficulties, we propose to deploy a digital twin of the production line by encoding its operational logic in a data-driven approach. By iteratively mapping the real-world data reflecting equipment operation status and product quality indicators in the digital twin, we adopt a quality prediction model for production process based on self-attention-enabled temporal convolutional neural networks. This model enables the data-driven state evolution of the digital twin. The digital twin takes a role of aggregating the information of actual operating conditions and the results of quality-sensitive analysis, which facilitates the optimization of process production with virtual-reality evolution. Leveraging the digital twin  as an information-flow carrier, we extract temporal features from key process indicators and establish a production process quality prediction model based on the proposed deep neural network. Our operation experiments on a specific  tobacco shredding line demonstrate that the proposed digital twin-based production process optimization method  fosters seamless integration between virtual and real production lines. This integration achieves an average operating status prediction accuracy of over 98\% and a product quality acceptance rate of over 96\%.
\end{abstract}


\section{Introduction}
The process industry serves as a core component across diverse industrial sectors including chemicals, food, pharmaceuticals and petroleum~\cite{YANG20211224}. The operation of process production lines directly impacts product quality and production efficiency, consequently influencing enterprise profits~\cite{Oztemel2020}. Nevertheless, ensuring long-term stability and optimal operation of production processes poses a significant challenge, given the involvement of multiple intricate production stages, equipment units, and numerous production parameters. Recently, emerging  concepts such as intelligent manufacturing, big data-driven state analysis and digital twins have been integrated into the management of the process industry~\cite{YAO20181, LU2020101837}. These advancements offer novel approaches to guarantee optimized long-term operation, specifically related to production quality in the process industry. In light of these developments, conducting research in this domain becomes crucial as it sheds light on new perspectives and solutions for production process optimization, ultimately helping industrial sectors improve product quality while minimizing production costs.

Unlike traditional discrete manufacturing~\cite{STOCK2016536}, process manufacturing involves a complex sequence of functional stages, where production control often faces a challenging dynamic optimization problem due to the coupling of parameter between processes and conflicts among multiple objectives~\cite{Jiongming2018ASO}. These complexities significantly hinder efforts to further improve production quality and resource utilization efficiency in process manufacturing. Fortunately, the emergence  of intelligent, data-driven optimization, as a novel decision-making approach, has demonstrated its effectiveness in handling large datasets and facilitating swift, accurate decisions~\cite{9723472}. By integrating data-driven intelligence into manufacturing management, the key to manufacturing process optimization now lies in real-time monitoring of various changes in both production process and product quality, followed by properly coordinating/optimizing all production stages through possibly high-dimensional feedbacks.

Process data in production lines typically manifests as a series of real-time, high-dimensional, and periodic data streams originating from multiple sources. These data streams encode the temporal-spatial interactions among various elements in the production process, and thus can be utilized to model parameter coupling in the industry process~\cite{Klingenberg2019}. In this study, we integrate different elements of the production process, including materials, equipment, process parameters, and process requirements, into a closely coordinated data-driven model. Our approach relies on the deployment of a digital twin, through facilitating information exchange between the physical entity of a process production line and its virtual counterpart~\cite{LU2020101837}. By doing so, we propose a deep learning model for production process state prediction, which is trained by the historical data collected from the physical side. The prediction model is used by the production optimization module for search of optimal operational parameters, with production process state updated and evaluated on the twin side. This approach enables the parallel evolution of the virtual and the physical production line. By feeding back the parameter solution from the digital twin to the physical process, we are able to improve the physical production line operation by predicting the production quality and recommending optimized process parameters without risking the real product quality. The main contributions of our work are as follows.
\begin{itemize}
    \item [1)] We propose an integrated framework for real-time state prediction and optimization of process parameters based on digital twins. The digital twin enables collaboration  between virtual and physical production lines, thus making it possible for operation optimization of the physical lines through iterative feed-ins from the digital twin side.
    \item [2)] We integrate the task of production line parameter adjustment into the process of real-time state monitoring and product quality prediction, and propose an evolutionary method-based parameter optimization approach based on digital twins to address the tradeoff between quality control and decision efficiency.
    \item [3)] In this study, we deploy a real digital twin system on a test production line for tobacco shredding. We provide a full sketch of the framework design for our twin system, which demonstrates a promising approach for integrating end sensors and actuators in the IoT network, the Manufacturing Execution System (MES), the information processing middleware and the AI functionalities. Our experimental study on the test line shows that the proposed digital twin framework can significantly improve production efficiency and, therefore, can serve as a prototype for applications in other industrial sectors.
\end{itemize}

The rest of the paper is organized as follows. Section~\ref{sec_relate_work} presents a brief summary of the related works. In Section~\ref{lab_DT_construction}, we present the proposed digital twin production line framework and highlight key components for the implementation. Section~\ref{Sec:method} introduces a real-time product quality prediction mechanism based on a sparse-attention-enabled deep neural network. Section~\ref{sec:Optimization} then proposes a heuristic parameter optimization method for the production line, which is built upon the line state prediction using the proposed neural network model. After that, Section~\ref{experiments} describes system implementation details and Section~\ref{sec:Results} validates the efficiency of our proposed framework. Finally, Section~\ref{sec:Final} concludes the paper.

\section{Related Work}
\label{sec_relate_work}

\subsection{Data-driven Quality Prediction and Control in Process Industries}
Leveraging Industrial Internet of Things (IIoT) networks, the evolution of industrial Internet technology has enabled process manufacturing enterprises to gather substantial historical data on production process characteristics, equipment operation, and quality indicators~\cite{YAO20181}. This makes it possible for the researchers to conduct various data-driven studies on quality prediction and process parameter optimization. For process quality prediction, a self-adjusting structural radial basis function neural network is proposed in~\cite{PMID:30986722} to address the difficulty in predicting online the outlet ferrous ion concentration for wet zinc smelting plants. In~\cite{Gongzhuang2022}, lower/upper-bound estimation is used to obtain the prediction interval for mechanical performance of hot rolled strip, and the weight parameters of the learning system is optimized using the artificial bee colony algorithm to predict the mechanical performance of hot rolled steel. In~\cite{9088162}, a quality prediction framework for process industries based on IoT-oriented cyber-physical system is proposed. In this way, key process indicators are selected through information gain and the analysis of sensitivity parameters is incorporated into a Bayesian optimization scheme integrated with random forests. Regarding a real industrial hydrocracking process, Chen et al.~\cite{9716776} proposed to utilize a regularized stacked autoencoder based on soft sensors to characterize the key process parameters.

For data-driven production parameter control, a multistage model is constructed to predict process parameters and quality indicators in~\cite{YIN2020106284} through two different connection strategies. Then, a multi-gene genetic programming and multi-objective particle swarm optimization algorithm is proposed to address the multi-stage and latency issues in the optimization of process manufacturing. A case study of coal preparation demonstrates the efficiency of the proposed method. With a similar goal, the research in~\cite{HU2019232} focuses on carbon efficiency in the iron ore sintering process. The comprehensive carbon ratio is predicted by integrating a fuzzy clustering method, a least-square support vector machine, and the Takagi-Sugeno fusion scheme. Based on the carbon efficiency prediction model, an online searching scheme using chaos particle swarm optimization is implemented to optimize the carbon ratio.

Despite the development of quality prediction and control techniques, the above-mentioned research in complex process manufacturing sectors still encounters subsequent obstacles. Firstly, quality forecasting and management of production lines typically rely on training with limited-size offline data, leading to potential accuracy degradation over long-term operation. Secondly, in the data-driven quality prediction/control approach of the process industry, the lack of interactive feedback from online data updates poses a challenge in achieving a systematic enhancement of operation quality in alignment with the real-time production line state.

\subsection{Digital Twins in Process Industries}
The concept of Digital Twins (DTs) involves establishing a mutual-mapping, with full information exchange, between the physical entity of a process production line and its virtual counterpart~\cite{LU2020101837}. DT has been considered a revolutionary opportunity in the digital transformation of process industries~\cite{8258937}.  The current research on DTs are primarily focused on their application in areas such as data management, fault diagnosis and prediction, dynamic scheduling in workshops, and  quality prediction for discrete manufacturing products. In~\cite{LIU2022857}, a cloud-edge collaboration framework is proposed to manage the full life cycle data of metal additive manufacturing. With the proposed framework, efficient data communication is guaranteed between field-level manufacturing equipment, edge twin bodies, and cloud twin bodies, thus facilitating cloud-based and deep learning-enabled defect analysis. In~\cite{ZHANG2021146},   data from physical and virtual entities are integrated, and the DT is utilized to fuse both real and simulated data and offer more comprehensive information about machine availability prediction and disturbance detection.

With DT, applications such as Prognostics and Health Management (PHM) are able to leverage generative data on the virtual DT side to enhance the accuracy of prognosis and decisions, especially when techniques demanding a large amount of data, such as deep learning, are employed. In~\cite{TAO2018169}, a DT-driven PHM system for complex equipment is proposed. A 5-dimensional DT (physical system, virtual equipment, service, data and connection) of a wind turbine generator is built by modeling the transmission relations among the wind wheel, gearbox, and generator. By fusing vibration and stress signals for fault prediction, a DT-based, simulation-driven fault prediction scheme is established. In~\cite{deebak2022digital}, a DT-assisted fault diagnosis method is proposed using deep transfer learning, analyzing the operating states of the machining tools. In~\cite{9762013}, a deep learning-based method for die-casting operation status analysis and appearance defect prediction is proposed. By establishing a virtual die-casting module, a joint virtual-real debugging process for the controller joint is introduced. With data acquisition using DT, two learning methods, i.e., XGBoost and a VGG16-based deep neural network, are adopted for quality prediction and appearance defect prediction, respectively.

In the pursuit of intelligent manufacturing systems, integrating Deep Reinforcement Learning (DRL) with DTs has also been considered an effective approach in enhancing decision-making quality~\cite{MULLERZHANG2023103933, PIRES2023103884}. Notably, research in this area highlights the benefits of DRL, especially in resource allocation and predictive maintenance applications~\cite{9714509, 9815106}. Furthermore, the implementation of DRL within DT has been explored for job shop scheduling in smart manufacturing~\cite{serrano2024job}. These investigations offer promising insights into the feasibility of efficient and flexible smart manufacturing practices. However, in complex manufacturing scenarios, these methods may suffer from instability and the lack of explainability.

Although significant efforts have been made to enhance data acquisition with DTs for tasks such as visualization and fault detection, there is still a lack of research focused on optimizing process parameters in production settings that involve temporal coupling among various sub-processes. Also, there is a need for finding efficient and explainable operation updating mechanisms for DT production lines, which are expected to integrate quality prediction and production control into the virtual product line evolution. Furthermore, the methods are expected to be able to self-evolve based on the interaction between the DT and the physical production line.
\begin{figure*}[t]
    \centering
    \includegraphics[width=0.8\linewidth]{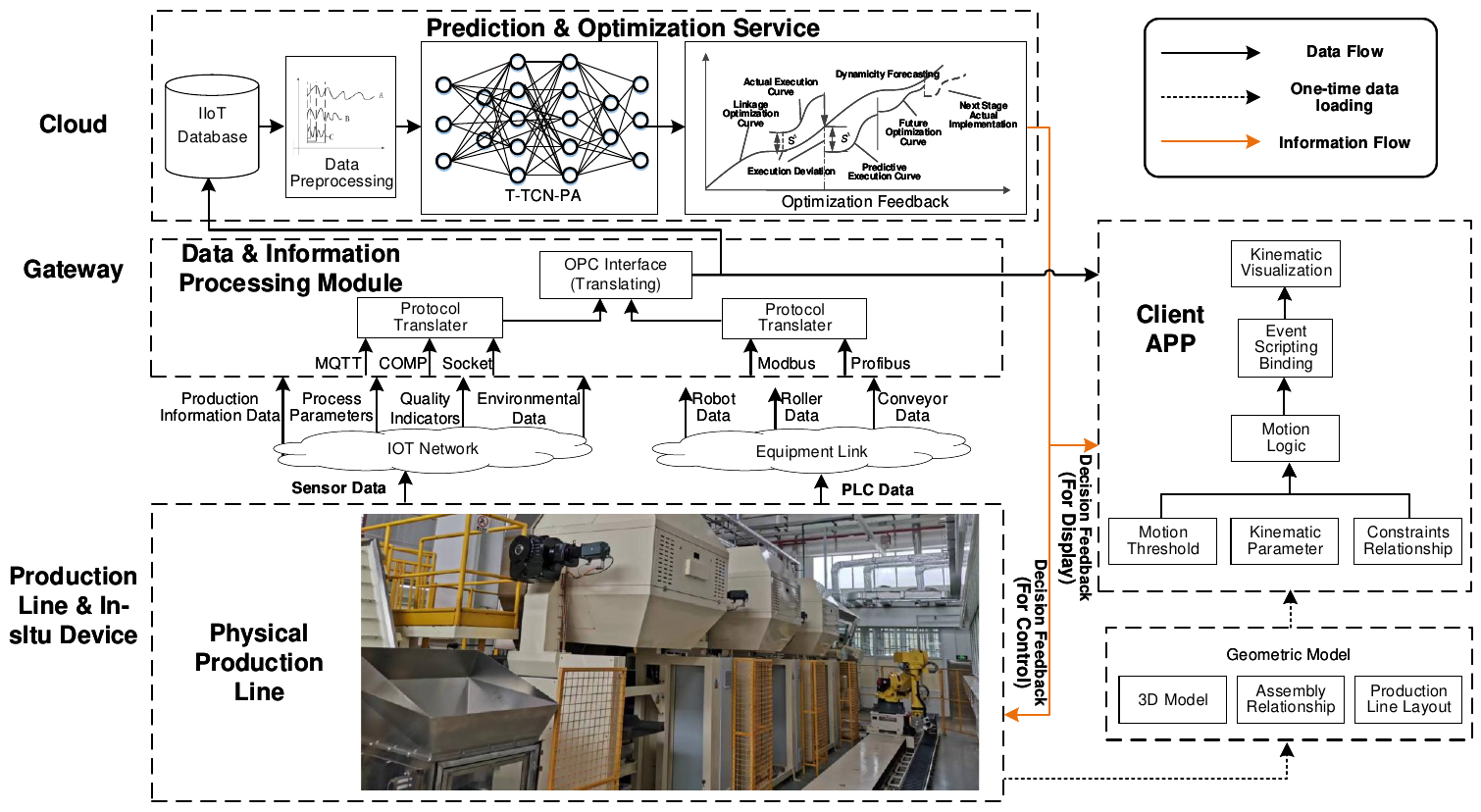}
    \caption{\small{The framework of DT for process production optimization. The system is composed of 4 major modules: the physical entities of the production line, the data and information processing module, the prediction and optimization service, and the client application. }}
    \label{fig:1}
\end{figure*}

\section{Construction of DT Model for Process Production Lines}
\label{lab_DT_construction}
Due to the complexity in physical processes and the kinematic relationships of machine parts, establishing precise mathematical models is often difficult, as traditional analytical methods struggle with highly nonlinear and coupled production systems of high-dimensional states. To address this issue, this study proposes a data-driven approach that uses deep learning models' black-box mapping capability to capture complex relationships between data inputs and outputs in the DT, instead of establishing explicit functional models.

\subsection{The Proposed DT Framework for Production Lines}
The deployment of a DT in process industries for quality prediction and optimization of line parameters involves the following key steps. Firstly, equipment and process parameters are collected offline from various equipment units to provide the necessary data for the construction of the DT model, which is used for both 3D modeling and the parameter design of deep learning models. This step establishes the mapping between the physical equipment entities and the DT. We note that it also converts the coupling between physical sub-processes into the casual relationship between the inputs and outputs of the DT model. Secondly, online monitoring of process states and quality fluctuations is conducted in real time. In the proposed framework, data are generated from different sources, such as wireless sensor readings from IIoT and PLC states from industrial buses. Data collection for the DT is achieved by using the OLE for Process Control (OPC) Unified Architecture (UA) protocol on dedicated IoT gateways that bridge different underlying sub-networks (e.g., IP/non-IP based IIoT, Profinet over serial ports and Ethernet). For instance, in an IIoT-enabled sub-network, sensors compatible with 802.11 may transmit production monitoring data using the Message Queueing Telemetry Transport (MQTT) protocol. A gateway relays the corresponding topics (sensor data of interest) as the MQTT broker, formats the received data into OPC-compatible packets, and subsequently transmits them to the DT host for further processing. If the data are collected from the Profibus-supported sub-network, we can directly use means such as the Socket-based interface or the KepServer software to conduct data conversion to the OPC UA server.

The data collected online from the production line describe the equipment movement and are partly used to power the digital 3D models for virtualization at the client end. The remaining data, which include process data and production parameters, are utilized by the algorithmic model to generate appropriate operational parameters. Using the process data aggregated at the DT side, we design a deep Neural Network (NN) model for predicting process quality indicators. A metaheuristic optimization procedure is performed with respect to key process parameters, guiding them toward the suitable ones that pass the evaluation by the NN. The parameter adjustment results are then fed back to the physical production line to improve the production efficiency and enhance the product quality. In Figure~\ref{fig:1}, a framework overview is presented to illustrate the key modules of the DT for a general process production line. This framework highlights the functionalities of the DT as a data aggregator as well as the host of the services for quality prediction and parameter optimization. More specifically, the DT framework presented in Figure~\ref{fig:1} comprises four main parts:

\begin{itemize}
    \item [(1)] \textbf{Physical Production Line:} This module includes two parts: physical equipment units and production data collection. Proper description of the geometric structure, such as the equipment structure, assembly relationships, and associated layout of the production line, is collected offline to construct the geometric model for visualization. Online production data including operational state data and quality indicators are collected in real time by heterogeneous in-situ sensors ~\cite{9585062}. The online data may be transmitted in different sub-networks, for example, the IoT network of wireless sensors and the Manufacturing Execution System (MES) deployed on the local area network. The heterogeneous production data are streamed to the IoT gateway for further processing.       
    \item [(2)] \textbf{Data and Information Processing Module:} This module is deployed in the IoT gateway at the edge, utilizing the OPC UA protocol to aggregate control signals and various sensor data in real-time from heterogeneous sub-networks in the field. The gateway subscribes to messages transmitted with different protocols such as MQTT, TCP, Modbus and Profibus, and aggregate them by OPC UA service for further forwarding. The operational data associated to the control logic of the production line may be directly sent to the user client, which utilizes these data as event triggers to synchronize the motion of the DT production line for visualization. The production state data are transmitted to the cloud database to support the training and inference processes in the intelligent decision module. The design of data flow will be detailed in the next subsection.

    \item [(3)] \textbf{Prediction \& Optimization Service:} Deployed on the cloud, this service provides plug-in encapsulation of a deep NN model for production quality prediction, along with a metaheuristic production optimization algorithm. The prediction model is trained using the dataset retrieved from the cloud database, and is utilized by the optimization algorithm for parameter fitness evaluation. The solutions of the optimization algorithm will be fed back to the physical entity for operational parameter updating.
    
    \item [(4)] \textbf{Client Application:}  The client application provides the human-system interface and is particularly responsible for the kinematic representation and visualization of the physical production line. It functions as the sink for DT data within the system, presenting process data in a human-comprehensible manner to assist engineers in making decisions regarding line operation. Frequently, the client app is used to refer to the entire twin system. Visualization of the physical production line is achieved based on the offline kinematic model construction, which will be further explained in Section~\ref{sub_3Dmodel_construction}.
\end{itemize}

\begin{figure*}[t]
	\centering
	\includegraphics[width=0.8\linewidth]{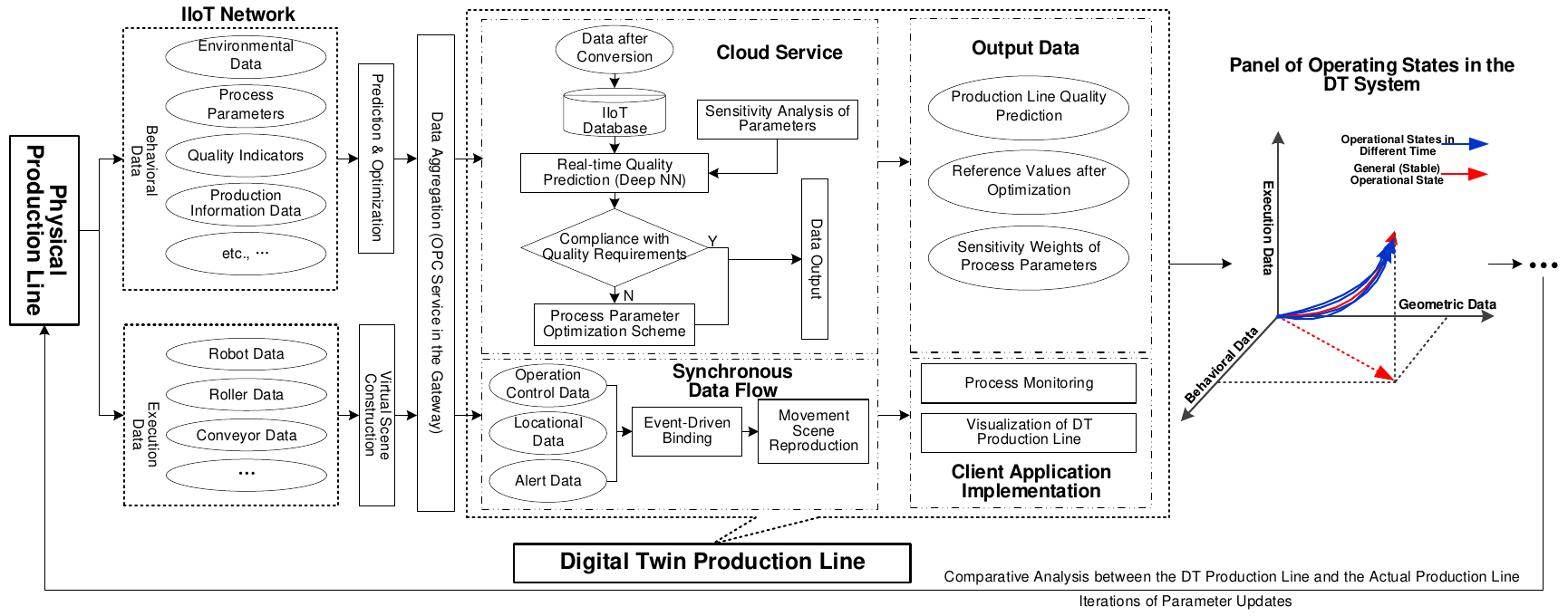}
	\caption{\small{Data processing flow of the DT production line.}}
	\label{fig:2}
\end{figure*}

\subsection{Data Flow in the DT Framework}
The proposed DT framework not only provides a kinematic representation of the production line in the visualization module, but also offers a dynamics model of the production process to reflect the complex physicochemical reactions that occur during production (see prediction \& optimization service in Figure~\ref{fig:1}). This design requires different processing flows of production process data and equipment data, which are dispatched by the data \& information processing module. 

As shown in Figure~\ref{fig:2}, data collected from the physical production line can be divided into two categories: offline data, which is used for virtual scene construction, and online data, which is used to power the 3D model operation as well as provide prediction and optimization services. The online data about the production line states is received from the Industrial IoT (IIoT) and then stored in a cloud database for later processing. On the other hand, online motion description data, including operation control data, location data, and mechanical alert data, are extracted from the MES system and then directly sent to the client app for visualization purposes. The optimization service deployed in the cloud uses the data retrieved from the IIoT database for product quality prediction and operational parameter optimization. The prediction outputs are then sent to the client app to update the operating state display panel. The optimization solution is sent to the physical production line for operational parameter updates.

\section{Quality Prediction and Parameter Optimization based on DT}
\label{Sec:method}

\subsection{Quality Prediction based on DT}
In practical industrial scenarios, due to the complexity of process production, time series data of production processes typically display characteristics such as nonlinearity, non-stationarity as well as long-term dependencies. This causes difficulties in identifying the causal relationships between the sensed production line states and the product quality. Meanwhile, due to the high data dimensionality, it is usually difficulty to locate the source of noises and identify outliers, which may lead to inaccurate analysis results and misleading predictions about the production process.
\begin{figure*}[t]
    \centering
    \includegraphics[width=0.7\linewidth]{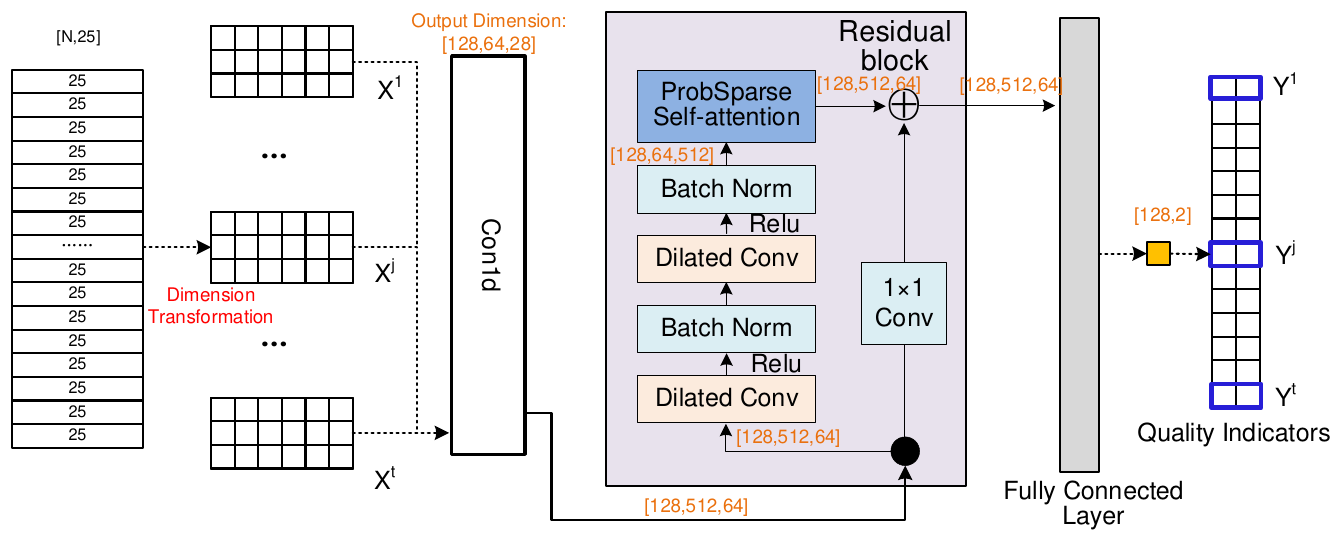}
    \caption{\small{The overview of the proposed NN model. It is composed of three major parts: input representation, stacked TCN with ProbSparse self-attention, and prediction head. The feature dimensions are determined according to the data of the production line considered in Section~\ref{experiments}.}}
    \label{fig:3}
\end{figure*}

In view of the above-mentioned issues, we propose an attention-based deep NN model for quality prediction to address the problems of nonlinearity and long-term dependencies in our obtained process data. The quality prediction module is located in the DT service module (see Figure~\ref{fig:1}). It receives data from the information processing middleware, utilizing the historical data retrieved from the IoT database for model training and the real-time data for inferences. To address the issue of non-stationarity in the original data, we design a data cleaning scheme to pre-process the collected sensor data and handle the scenarios of machine stoppage and feedstock fluctuations at warm-up and downtime of the production line. This removes most of the noise and outliers in the collected sensor data while ensuring the integrity of the data. Using the data set obtained after pre-processing, we aim to address the long-term dependencies in the multidimensional sequential data in the product-quality forecasting task. The proposed NN model is built by incorporating the Probabilistic Sparse self-Attention (PSA) mechanism~\cite{zhou2021informer} into the Time Convolutional Network (TCN) framework~\cite{9481153}. The reason for introducing the self-attention mechanism lies in the fact that traditional TCNs rely on changing the number of output channels to extract temporal dependence feature. Hence, its capability of extracting the correlations between multiple variables of production process parameters is insufficient. Therefore, sparse self-attention is employed to captures such correlations with a reduced computational load. The structure of the network is composed of three main components, that is, the input representation layer, the attention-enabled TCN layer and the prediction head (see Figure~\ref{fig:3}).
\begin{figure*}[t]
    \centering
    \includegraphics[width=0.7\linewidth]{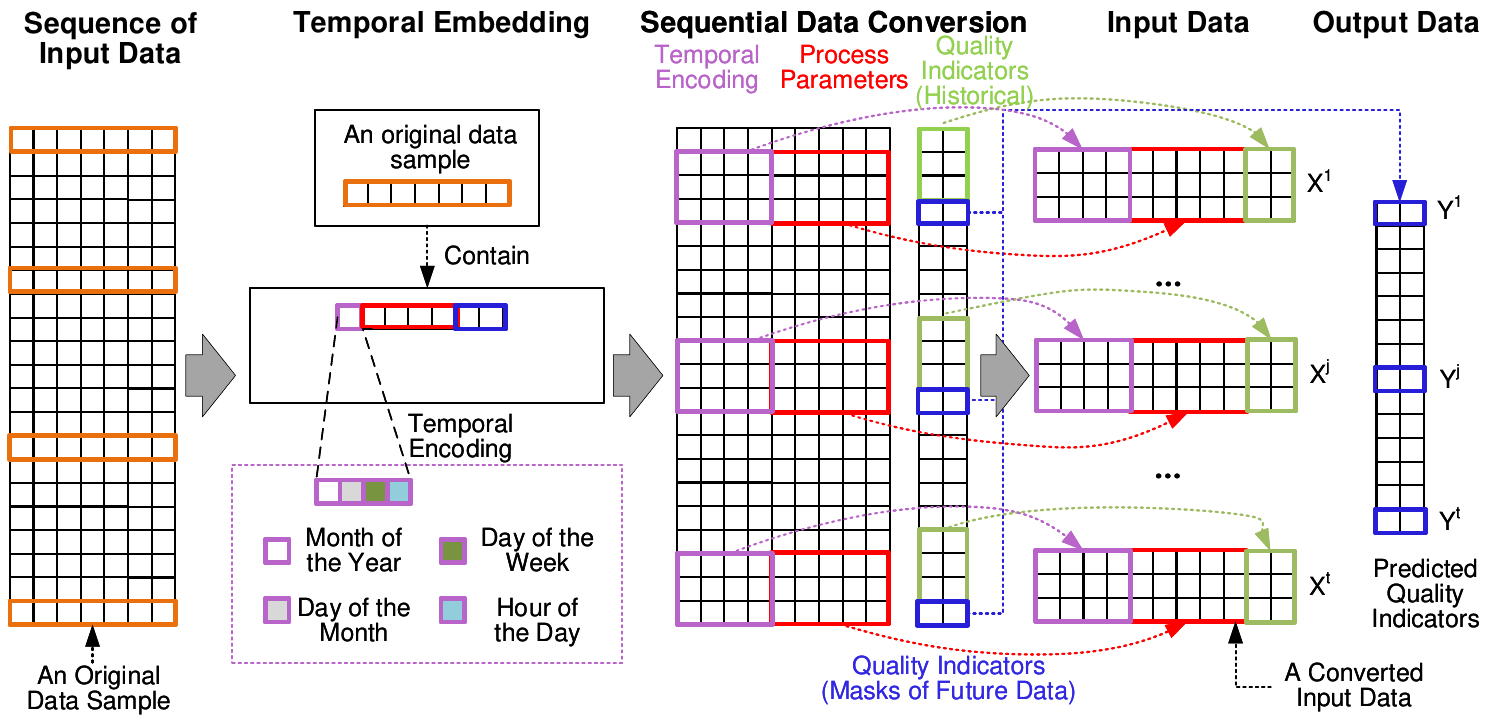}
    \caption{\small{Re-organization of the raw input data sequence, where different colors represent different types of data chunks.}}
    \label{fig:4}
\end{figure*}

\subsection{Input Representation}
We observe that different stages of the production with batch processing of inlet material often exhibit periodic characteristics. To preserve the correlated information between stages more effectively, the original input data, presented as a sequence of vectors, is converted into a consistent format of sequential matrix data through sequence merging and data normalization. As illustrated in Figure~\ref{fig:4}, an original data vector contains its associated timestamp information, process parameters, and corresponding quality indicator records. At the input representation layer, the original timestamp of each data sample is encoded from a scalar format to a four-element vector format, with each column encoding the time information in different granularity, i.e, the month, day, week, and hour, respectively. The primary goal of timestamp conversion is to provide a structured data representation for time series analysis, enabling neural networks to effectively learn and analyze time-dependent patterns.

Subsequently, for a specific time instance $t$, the process parameters are denoted as an $m$-dimensional vector $x^t=[x^t_1,\ldots, x^t_m]$, and the quality metric data is expressed as an $n$-dimensional vector $y^t=[y^t_1,\ldots, y^t_n]$. Given a fixed-length time window of $T$ samples sequentially, the input data sample $X^t$ for the NN model is constructed by concatenating $T$ sequential samples retrieved from the database till time instance $t$, i.e.,  $(x^{t-T+1}, \ldots, x^t)$ and the $T$ historical quality indicators by shifting their corresponding quality indicators for one time step left, i.e. $(y^{t-T}, \ldots, y^{t-1} )$. The corresponding label data $Y^t$ is taken as the quality indicator for the current time instance $y^t$. Formally, $X^t$ is in the following matrix form:
\begin{equation}
  \label{eq_X_t}
  X^t=\left[\begin{matrix}
    x^{t-T+1}_1 & \ldots & x^{t-T+1} &y^{t-T}_1 & \ldots & y^{t-T} \\
    \ldots & \ldots  & \ldots& \ldots  & \ldots \\
    x^t_1 & \ldots & x^t_m &y^{t-1}_1 & \ldots & y^{t-1}
  \end{matrix}\right].
\end{equation}

It should be noted that all the process parameters and quality indicators are normalized using the maximum-minimum normalization method into the range of $[0,1]$, which handles the potential problem of vast difference between the quantities of process parameters and those of the quality indicators obtained by heterogeneous sensors. Also, instead of directly adding the temporal embedding to the input data as in vanilla Transformer, we concatenate the normalized temporal encoding data (after broadcasting) to the front of $X^t$.

\subsection{TCN with Probabilistic Sparse Self-Attention}
\label{subsec_TCN}
\begin{figure}[t]
	\centering
	\includegraphics[width=1\linewidth]{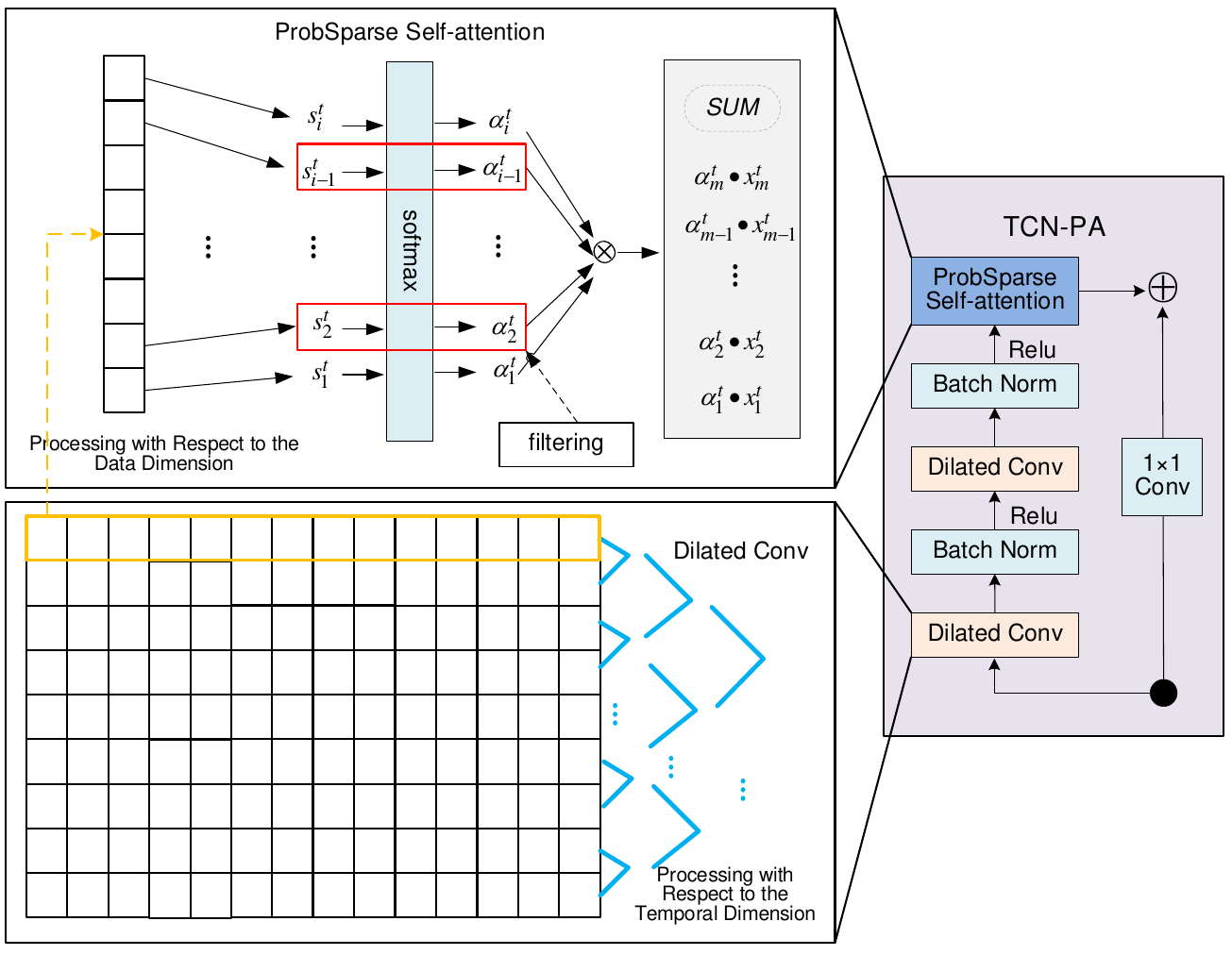}
	\caption{\small{The structure of the proposed NN model composed of TCN and PSA, where the attention score is computed based on (\ref{eq_qkv})-(\ref{eq_attention}).}}
	\label{fig:5}
\end{figure}

As illustrated in Figure~\ref{fig:3}, the re-organized input is fed into a 1D convolution layer to fit its dimension with the required dimension of the TCN layer. The input data is expanded from a dimension of $4+m+n$ (each element corresponding to the dimension of temporal encoding, production parameters, and quality indicators, respectively) into $N$, while the time window length for the input data samples is kept as $T$. This makes the input dimension of the TCN layer $T\times N$. The 1D convolution layer is connected to a layer of TCN with the Probabilistic sparse self-Attention (TCN-PA), which is composed of four modules including dilated convolution, batch normalization, PSA and identity mapping, as illustrated in Figure~\ref{fig:3} and further explained in Figure~\ref{fig:5}.

As shown in Figure~\ref{fig:5} (see bottom-left therein), by introducing a dilation factor $d$ in the dilated convolution module of the TCN, the convolution kernel skips a certain number of elements on the input sequence for convolution. The convolution operation is performed at certain intervals of the input time series. The computation of dilated convolution can be expressed as follows:
\begin{equation}
  \label{eq_dilated}
  H(t)=\sum_{k=0}^{K-1}X(t-d\times k) W(k),
\end{equation}
where $W$ contains the weights of the convolutional kernel of size $K$, $k$ is the index of the weight, $X$ is the input sequence, and $t$ is the time index. $d$ controls the operation spacing of each element in the convolution kernel within the input sequence.

The batch normalization layer is used to improve the robustness of the NN model. It performs normalization operations on the input data of each batch, ensuring that its output has a mean of 0 and a variance of 1. The operation of batch normalization can be represented as follows:
\begin{equation}
  \label{eq_bn}
  B(t)=\textrm{BN}(H(t)) = (H(t)-\mu)/\sigma,
\end{equation}
where $B(t)$ represents the result of batch normalization at time $t$, $H(t)$ is the output of dilated convolution following (\ref{eq_dilated}), and $\textrm{BN}(\cdot)$ represents the batch normalization operation. $\mu$ and $\sigma$ are the mean value and the standard deviation of the current batch input, respectively.

After stacking the operations of dilated convolution and batch normalization, which aim to expand the reception field, the output is used to generate the self-attention score. Furthermore, to reduce the computational cost of self-attention, we propose to adopt the probabilistic sparse self-attention mechanism~\cite{zhou2021informer} to improve the training efficiency. Similar to the traditional self-attention mechanism, the output of the final batch normalization layer in the TCN module undergoes linear transformation for 3 times with different learning weights to obtain the query matrix $Q$, key matrix $K$, and value matrix $V$, respectively, as follow:
\begin{align}
  \label{eq_qkv}
  Q=W_q B^{\textrm{T}}(t),\\
  K=W_k B^{\textrm{T}}(t),\\
  V=W_v B^{\textrm{T}}(t),
\end{align}
where  $W_q$, $W_k$, and $W_v$ are learning weights of the three transformation matrices. Let $q_i$, $k_i$ and $v_i$ denote the $i$-th column vectors of $Q$, $K$ and $V$, respectively. Then, we can compute the similarity $s_i^j$ between query $q_i$ and key $k_j$ to obtain a measurement of the correlation between different process feature vectors $i$ and $j$ as:
\begin{equation}
  \label{eq_similarity}
  s_i^j=q_i^Tk_j.
\end{equation}

Sparse self-attention uses the max-mean measurement in (\ref{eq_max_measurement}) to select the top-$M$ attention probabilities for query $q_i$ with the key vectors in $K$, and set the rest attention probabilities to $0$~\cite{zhou2021informer}. (\ref{eq_max_measurement}) is an approximation of the KL divergence, which theoretically measures the ``likeness'' (equivalently, the sparsity) of $q$'s and $k$'s.
\begin{equation}
  \label{eq_max_measurement}
  M(q_i, K)=\max_j\left\{\frac{s_i^j}{\sqrt{p}}\right\}-\frac{1}{L_k}\sum_{j=1}^{L_k}\frac{s_i^j}{\sqrt{p}},
\end{equation}
where $L_k$ represents the number of key vectors, and $p$ represents the dimensionality of the vectors. The $M$ non-zero similarities determines the attention probabilities with a softmax operation as follows:
\begin{equation}
  \label{eq_softmax}
  a_i^j=\textrm{softmax}(s_i^j)=\frac{\exp{(s_i^j)}}{\sum_{k=1}^{L_k}\exp{(s_k^j)}}.
\end{equation}
Using (\ref{eq_softmax}), the hidden states obtained from the TCN layer at each input produce the following context information state vector through weighted summation:
\begin{equation}
  \label{eq_attention}
  c_i=\sum_{i=1}^{L_k}a^j_i k^j_i.
\end{equation}

Finally, considering that process feature information is prone to loss or distortion after multiple layers of convolution and compression operations, a residual connection is introduced. This allows the network to retain information across layers, ensuring the preservation of the original data-dependent information. A residual block consists of an identity mapping component for the TCN output $X'(t)$ and a residual component for the input $X(t)$:
\begin{equation}
  \label{eq_residual}
  Z(t)=\textrm{1DConv}(X(t))+X'(t),
\end{equation}
where the 1D convolution operation $\textrm{1DConv}(\cdot)$ fits the dimension of the input data $X(t)$ with that of $X'(t)$. It can be computed as follows:
\begin{equation}
  \label{eq_1dconv}
  X^{\textrm{out}}_i(t)=\sigma\left(\sum_{k=0}^{K-1}W_{c,k}X_{i\times s+k}(t)+b\right),
\end{equation}
where $i$ indicates the element position of the input/output vector, $\sigma(\cdot)$ is a proper activation function, $W_{c,k}$ is the $k$-th element of the convolution kernel $W_c$, $s$ is the stride of the convolution operation, and $b$ represents the bias.

\subsection{Prediction Head}
The output of the TCN-PA module is connected to one or several fully connected layers to further manipulate the output dimensions, producing the final prediction of the quality indicators of concern (see Figure~\ref{fig:3}). We adopt the Mean Square Error (MSE) loss to train the proposed network.

\section{Production Line Optimization with Multi-objectives}
\label{sec:Optimization}
\subsection{Objective Function Design}
Intuitively, multiple quality indicators of the production process are directly influenced by the process operation parameters. With the proposed DT framework, our objective in process optimization is to minimize the deviation between the real-time values and the target values of the quality indicators, by controlling the operation parameters based on the feedback from the twin side. We take each quality indicator as a a single objective. Then, optimizing multiple quality indicators collectively forms a multi-objective optimization problem. Considering that the importance of each quality indicator varies, individual objective functions are weighted based on field experience. Additionally, given the significant differences in the fluctuation ranges of various quality indicators, normalization is applied to each single objective function. Omitting the time instance, the multi-objective optimization can be established as follows:
\begin{equation}
  \label{eq_opt}
  X^* =\arg\min_{X} \left\{U(X)=\sum_{i=1}^{n} l_i \left\vert \frac{f_i(X) - obj_i}{Y_i^{\max}-Y_i^{\min}}\right\vert  \right\},
\end{equation}
where $X$ is the input matrix obtained from the time series of process parameters with $t$ rows and $m$ columns, with $t$ representing the number of time slots and $m$ representing the dimension of input production parameters (see also Section~\ref{subsec_TCN}). $f(X)=[f_1(X),\ldots, f_n(X)]$ denotes the $n$-dimensional prediction output of the proposed NN model as described in Section~\ref{Sec:method}. $l_i$ is the weight of the $i$-th objective. $Y_i^{\max}$ and $Y_i^{\min}$ are the maximum and minimum values of the $i$-th quality indicator, respectively, which are recorded in the real-world production scenarios. 

We note that the optimization utilizes the input-output relationship embedded in the NN model on the DT side for objective function evaluation, while its solution to (\ref{eq_opt}) is implemented on the physical production line through information feedback to the various actuators. Therefore, the prediction accuracy of the NN model has a direct impact on the quality of operation control in the process production line.

\subsection{Extraction of Controllable Process Parameters}
\label{sub_sec_extraction}
Consider that $n$ process parameters are collected from the production line to form the parameter vector $X$ in (\ref{eq_opt}). In practice, not all of these parameters can be controlled. Generally, the $n$ process parameters can be divided into three categories. The first category includes environmental parameters that cannot be controlled or that change passively, such as ambient temperature and humidity. These parameters are determined by external environmental conditions and cannot be adjusted through our process control. The second category comprises non-independent parameters that are influenced by changes in the parameters of the third category in actual production. For example, the exhaust air volume and the main steam temperature of a drying machine are affected by other parameters. The third category consists of parameters that have a significant impact on quality indicators as well as the second-category parameters. Adjustments to them, such as the inlet air temperature and inlet air flow rate, can be quickly reflected in the production process outputs. The process parameters that we need to control belong to the third category. 

To effectively control the production process, we need to first determine the set of process parameters and then identify their categories based on the analysis of on-site operational conditions. Although the first-category parameters are important in determining the process operational states, they are not the primary focus of our adjustment algorithm due to the lack of controllability. After identifying the parameters of the first category, we employ the Sobol sensitivity analysis method~\cite{zhang2015sobol} to distinguish between the second and third categories of process parameters. The Sobol index quantifies the influence of input parameters on outputs through variance decomposition, categorizing those significantly affecting production quality as third-category parameters. Subsequently, the optimization of the production process is conducted with regard to these third-category parameters in $X$.

\subsection{Parameter Optimization using Deep NN-based Performance Evaluation }
The functional relationship $f(X)$ between the process parameters and the quality indicators is implicitly learned with the proposed deep NN model. Due to the lack of an explicit causal relationship model, we rely on meta-heuristic algorithms to search for the proper solution of the controllable parameters, using the inference results of the NN for evaluating the objective functions in (\ref{eq_opt}). In conventional swarm intelligence algorithms, it typically needs carefully hyperparameter design to balance exploration and exploitation of the searching agents in complex problems of different scales.
To address this issue, we introduce the Archimedes Optimization Algorithm (AOA)~\cite{9723472,2020Archimedes}  to improve search efficiency. With the controllable process parameters identified through sensitivity analysis, the values of the non-controllable or non-independent process parameters can be temporarily fixed at the current round of the optimal solution searching process.  With AoA, a random number of searching particles are generated within a specified range for search initialization. Since excessive fluctuations cannot occur during the processing of the products, we impose search limits on the value adaptation range of the controllable parameters in $X_t$. Empirically, assume that the fluctuation range of optimized process parameters cannot exceed $\epsilon$ ($0<\epsilon<1$, for which we set $\epsilon=1/8$) of their maximum fluctuation range. The upper and lower limits of the particle search range for the $i$-th parameter are given by
\begin{equation}
  \label{eq_lb_up}
  \left\{
  \begin{array}{ll}
    lb_i = x^{cur}_i -\displaystyle\epsilon(x^{\max}_i-x^{\min}_i),\\
    ub_i = x^{cur}_i +\displaystyle\epsilon(x^{\max}_i-x^{\min}_i),
  \end{array}
  \right.
\end{equation}
where $lb_i$ and $ub_i$ represent the lower bound and the upper bound of the search particle, respectively. $x^{cur}_i$ is the current value of the $i$-th process parameter. $x^{\max}_i$ and $x^{\min}_i$ are obtained from historical records as the maximum and minimum value of this parameter.

The evolution dynamics of the particles is defined by emulating the immersed objects with random volumes, densities and accelerations in the same fluid. The search trajectory of a particle is subject to buoyant force and collision from the other immersed objects. Particles attempt to attain the optimal process parameters corresponding to the optimal quality indicators in the current production state, where their emulated objects achieve a state of neutral buoyancy equilibrium. The particle positions are initialized as ($\forall i=1,\ldots, m', \forall k=1,\ldots, K$):
\begin{equation}
  \label{eq_particle_init}
    x_{i,k}=lb_i + {r}_{i,k} (ub_i-lb_i),
\end{equation}
where $x_{i,k}$ represents the initialized value of the $i$-th process parameter for the $k$-th search particle, and $lb_i$ and $ub_i$ are obtained with (\ref{eq_lb_up}). $m'$ denotes the number of controllable parameters, $K$ denotes the number of search particles,  and $r_{i,k}$ is a uniformly random number drawn in the range $[0,1]$.

In addition to the position in the search space, each particle needs to determine its density, volume, and acceleration to emulate an immersed object. Let $D_k$, $V_k$ and $A_k$ denote the density, volume, and acceleration of each particle object, respectively. $D_k$ and $V_k$ are also initialized with a uniformly random number, and $A_k$ is initialized similarly to in (\ref{eq_particle_init}) in a sclalar manner. The search/computation is performed based on the initial population, and the optimal fitness value is selected as the target to determine the optimal attributes of the searching particles $\mathbf{x}_{best}$, ${D}_{best}$, ${V}_{best}$ and ${A}_{best}$.

After initialization, the density and volume of each particle are updated as follow:
\begin{align}
    \label{eq_den_update}
  & D_k^{iter+1} = D_k^{iter} + r^{iter}_{k,D}\left({D}_{best}-D_k^{iter}\right),\\
  \label{eq_vol_update}
  & V_k^{iter+1} = V_k^{iter} + r^{iter}_{k,V}\left({V}_{best}-V_k^{iter}\right),
\end{align}
where ${D}_{best}$ and ${V}_{best}$ are the values of density and volume associated with the best particle at the current round, and $r^{iter}_{k,D}$ and $r^{iter}_{k,V}$ are the random values uniformly drawn from $[0,1]$.

During the particle optimization process, we simulate collision-incurred evolution rules between particles to control the exploration trajectory of each particle. After evolution of a certain period of time, the particle objects will reach an equilibrium state~\cite{2020Archimedes}, thus reducing unnecessary search. To simulate this scenario, we introduce the transfer factor $\alpha$ and the density decreasing factor $\delta$ to control the transition between exploration and exploitation. The two factors are iteratively updated as follows:
\begin{align}
    \label{eq_update_alpha}
  & \alpha^{iter+1} = \exp\left(\frac{iter - iter_{\max}}{iter_{\max}}\right),\\
  \label{eq_update_delta}
  & \delta^{iter+1}=\exp\left(\frac{iter - iter_{\max}}{iter_{\max}}\right)-\frac{iter}{iter_{\max}},
\end{align}
where $iter$ is the current iteration number, and  $iter_{\max}$ is the maximum number of iterations for particle evolution.

Given the values of $\alpha$,  the acceleration of the search particles is updated by iteration. The particles are set in the exploration phase when $\alpha^{iter}\le 0.5$, which simulates the condition of particle collisions. In this case, the particle acceleration for the $i$-th parameter is updated by simulating the collision with a random material:
\begin{equation}
  \label{eq_update}
  A_k^{iter+1} = \frac{D_{rand}V_{rand}A_{rand}}{D_k^{iter+1} V_k^{iter+1}},
\end{equation}
where $D_k^{iter+1}$ and $V_k^{iter+1}$ are obtained with  (\ref{eq_den_update}) and (\ref{eq_vol_update}), and $D_{rand}$, $V_{rand}$ and $A_{rand}$ are the randomly generated attributes of the colliding object.

When $\alpha^{iter}>0.5$, the particles develop and no collision occurs. Then, the particle acceleration is updated as
\begin{equation}
  \label{eq_update_develop}
  A_k^{iter+1} = \frac{D_{best} V_{best} A_{best}}{D_k^{iter+1}  V_k^{iter+1}},
\end{equation}
where ${D}_{best}$, ${V}_{best}$ and $A_{best}$ are the attributes of the best object so far.

Subsequently, the particle acceleration is normalized, with the parameters $\mu$ and $\eta$ limiting the normalization range ($\mu$ and $\eta$ are typically set to 0.9 and 0.1 according to~\cite{2020Archimedes}):
\begin{equation}
  \label{eq_update_normalize}
  {A}_{k, norm}^{{iter}+1} = \mu\frac{A_k^{iter+1}-{A}_{\min}}{{A}_{\max}-{A}_{\min}}+\eta,
\end{equation}
where ${A}_{\max}$ and ${A}_{\min}$ are the maximum and minimum allowable acceleration. ${A}_{k, {norm}}^{{iter}+1}$ determines the step size that each particle can change with. If particle $k$ is far from the global optimum, ${A}_{k, {norm}}^{{iter}+1}$ is higher, indicating that the particle is in the exploration phase. Otherwise, the particle is in the exploitation phase. (\ref{eq_update_normalize}) also impose a condition that the acceleration factor starts at a large value and decreases over time, determining the transition of particles from the exploration phase to the exploitation phase. This helps to prevent particles getting stuck in local optimal solutions.

With the updated attributes of density, volume and acceleration, the particle position is updated. Similarly, in the exploration state ($\alpha^{iter}\le 0.5$), we have for particle $k$:
\begin{align}
  \label{eq_pos_update}
  x_k^{iter+1} = x_k^{iter} +c_1 r_k^{iter} A_{k,norm}^{iter+1} \delta^{iter}(x^{iter}_{rand}-x_k^{iter}),
\end{align}
where $c_1$ is a constant (heuristically, $c_1=2$), $r_k^{iter}$ is a randomly generated number for particle $k$. $x^{iter}_{rand}$ is the position of a randomly selected particle among the $K$ particles. $\delta^{iter}$ is obtained with  (\ref{eq_update_delta}).

In the exploitation state ($\alpha^{iter}> 0.5$), we have 
\begin{align}
  \label{eq_pos_update_developed}
  x_k^{iter+1}
  = x_{k,best}^{iter} +\theta c_2 r_{k}^{iter} A_{k,norm}^{iter+1}  \delta^{iter}(\tau x_{best}-x_k^{iter}),
\end{align}
where $c_2$ is a constant (heuristically, $c_2=6$), and $r_{i}^{iter}$ is a randomly generated number as in (\ref{eq_pos_update}). $\theta$ sets whether to change the searching direction:
\begin{equation}
  \label{eq_F}
  \theta=\left\{
  \begin{array}{ll}
    +1, {\textrm{ if }} 2r_{\theta}-c_4\le 0.5,\\
    -1, {\textrm{ otherwise}},
  \end{array}
  \right.
\end{equation}
where $r_{\theta}$ is a random value again and $c_4$ is a constant. Finally, $\tau$ increases with time, and determines a certain searching position that is proportional to the current best object. With $c_3\in[0,1]$ we confine the value of $\tau$ as
\begin{equation}
  \label{eq_T}
\tau=\max(c_3 \alpha^{iter},1),
\end{equation}
where $c_3 \alpha^{iter}$ indicates that $\tau$ is directly proportional to the transfer factor $\alpha$.

\begin{figure}[t]
    \centering
    \includegraphics[width=0.95\linewidth]{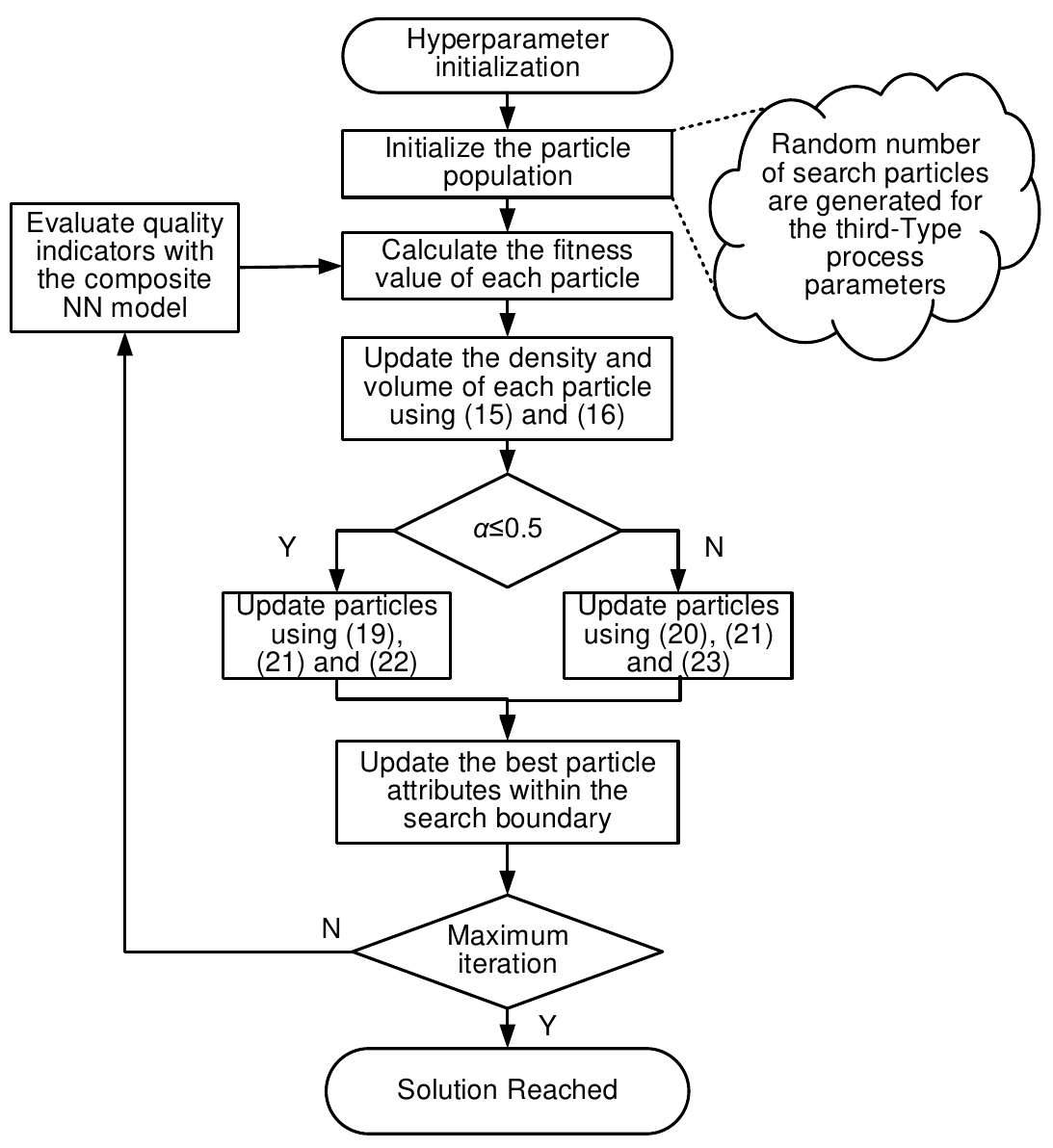}
    \caption{\small{ Schematic of the optimization scheme based on AOA.}}
    \label{fig:7}
\end{figure}

We note that by introducing $\tau$ and $\theta$, a larger step size of random walks is generated in (\ref{eq_pos_update_developed}) at the beginning of the particle evolution. As the search progresses, this step size gradually decreases, hence the difference between the best position and the current position gradually decreases. The step size of the particle position evolves from large to small, ensuring that the particles find the best position as quickly as possible. The operation flow of AOA is shown in Figure~\ref{fig:7}.

\section{System Deployment}
\label{experiments}
\subsection{Deployment of DT on an Experimental Line}
To validate the effectiveness of our DT-based optimization framework, a tobacco shredding line, with a processing capacity of 20 kg/batch, was selected for a case study of real-time process monitoring and control. Based on the established data collection and processing logic (see Figures~\ref{fig:1} and~\ref{fig:2}) for key indicators of the tobacco leaf drying process, specifically, the ``thin-plate drying'' process, the controlled product quality was evaluated. The proposed composite NN is used for quality prediction. If the predicted quality indicators fell below the expected levels, the proposed optimization algorithm was activated to adjust the operating parameters on the DT side. The updated parameters were then sent from the DT side to the physical production line. Sensor Data were collected every 6 seconds on the test production line.

\begin{figure*}[t]
    \centering
    \includegraphics[width=0.8\linewidth]{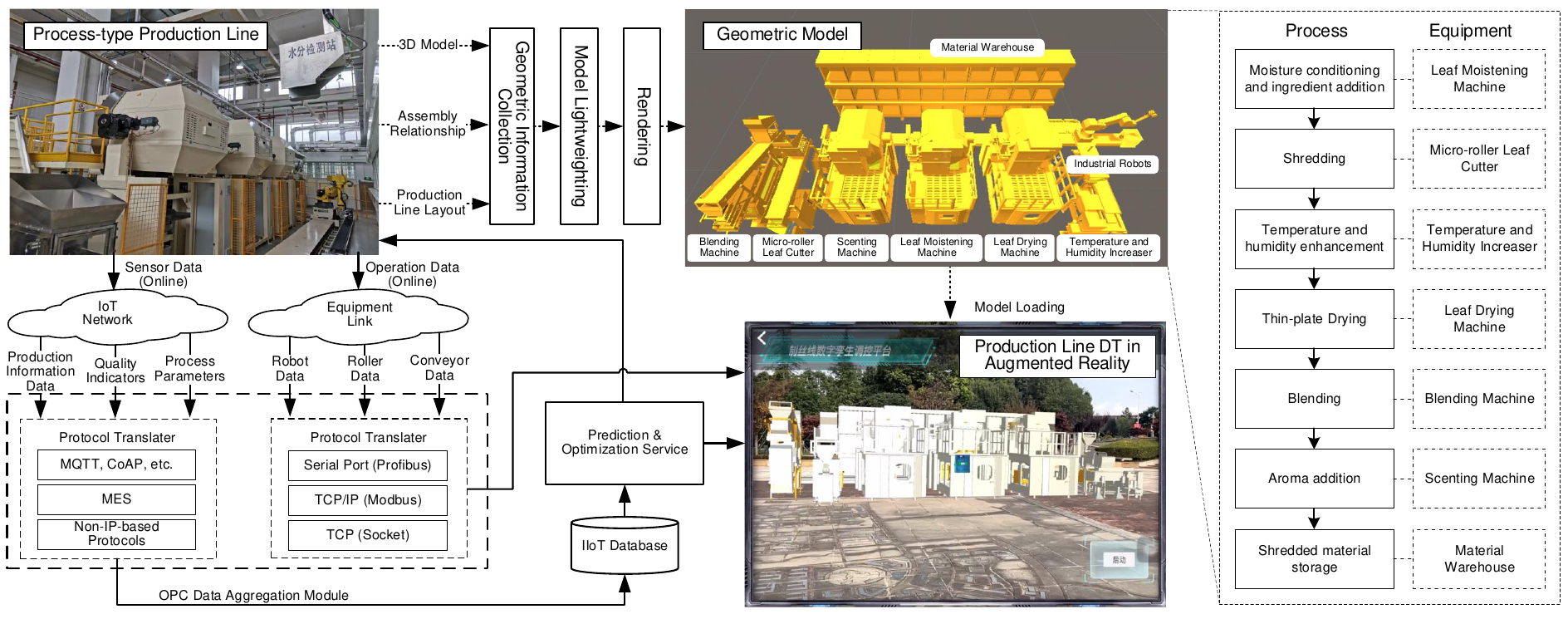}
    \caption{\small{The procedures for constructing the 3D model of a tobacco shredding production line in the DT system.}}
    \label{fig:8}
\end{figure*}

Our system is deployed based on the following equipment units: the blending machine, the micro-roller leaf cutter, the scenting machine, the leaf moistening machine, the leaf drying machine, the temperature and humidity increaser, and the industrial robots. Each machine corresponds to a working procedure, with the robot moving the material container between different machines (see Figure~\ref{fig:8}). The DT software is developed based on the following tool chain: Kepware for data collection, unity3D for visualization, MySQL for data storage, and Pytorch for NN model deployment. The details of the system deployment are organized in the following steps:
\begin{itemize}
  \item[(1)] At the IoT gateway/edge, we deploy a Kepware-based real-time data acquisition service, translating the data of different protocols into OPC-UA compatible data stream. The key data, such as the moisture of inlet materials, opening levels of the exhaust damper, steam pressure into the thin plate, etc., are collected by multiple sensors and transmitted with different protocols such as MQTT, TCP and Profibus to the gateway. After data translation, real-time sensor data are pushed to the cloud database. The operational parameters associated  to each batch of inlet material are also stored in the database at the beginning of the processing batch.
  \item[(2)] In the cloud, the sensor data for the production line are further processed before storing into the database. This is mainly about preprocessing invalid data, including outlier treatment, denoising, identification of the beginning and stoppage of material inlets, and data normalization.
  \item[(3)] The proposed composite NN model is deployed as part of the prediction and optimization service in the cloud. It is trained offline using the dataset obtained with Step (2). After training is completed, the NN is turned into online inference mode for real-time product quality prediction. Its inference output is used to invoke the AOA algorithm in the same service to evaluate the search results of production process parameters, of which the best result is fed back to the physical production line.
  \item[(4)] The DT data obtained in Steps (2) and (3) are communicated to the visualization module in the client application through Socket tunnel, with Unity3D serving as the rendering engine. The XCharts plugin is used to generate the panel for online operational status displaying, as well as visualization of prediction and control results.
  \item [(5)] Finally, the physical production parameters are adjusted with PLCs according to the optimization service feedback. This completes the four-phase DT-to-physical-entity update through data collection, visualization, parameter prediction, and process optimization. The updated data due to physical adjustment are stored into the on-cloud database and are used to expand the training dataset for transfer learning of the prediction model.
\end{itemize}

\subsection{Data Collection and Pre-processing}
\begin{table}[t]
  \centering
  \caption{Rules for identifying non-steady-state samples according the state of material flows.}
\begin{tabular}{p{2.5cm}| p{2.5cm}| p{2.5cm}}
\hline
 Determinant Factor & Conditions for Identifying the Valid Data & Conditions for Stopping Recording the Valid Data \\ \hline
Accumulated material volume & The first sample with unchanging volume  & The last sample with unchanging volume \\ \hline
 Material flow rate $r$ &  $r<30$ & $r\ge 30$ \\ \hline
Inlet beginning and stoppage & Synchronized with the beginning of moisture contents at the outlet & Synchronized with the stoppage of material flows at the inlet\\
 \hline
\end{tabular}
\label{Table:1}
\end{table}

The data used in this paper were acquired from the tobacco shredding test line of an anonymous process manufacturing enterprise. This line comprises six processes including humidity conditioning, primary feeding, secondary feeding, thin-plate drying, proportional material blending, and flavoring. The dataset used for training the prediction module includes over 200,000 records from 45 batches produced between June and December 2023.

As mentioned earlier in this paper, process manufacturing is characterized by continuous production. A key aspect of preprocessing the production data is cleaning the head and tail data, as well as the data collected from line stoppages and breaks. The rules for identifying non-steady state data at the start and end of production are outlined in Table~\ref{Table:1}. The data truncation is enabled by the following three conditions: (1) if there is no increase in the measurement of ``accumulated material volume'' within the data segment; (2) if the detected material flow rate is less than 30; (3) if the first data sample in the segment synchronizes with the beginning of the moisture contents at the process outlet.

\subsection{Construction of Visualization Module}
\label{sub_3Dmodel_construction}
We use 3D laser scanners to collect offline the geometric information of the production line, such as the size and shape of equipment parts, equipment structure, assembly relationships of machines, and production line layout. Then, the collected geometric data are processed with the UG modeling software to create scale models of parts, equipment units, and the production line. Based on these data, a high-precision 3D model of the production line is established. We use 3ds MAX to add textures to the geometric models, which are required by the client application for rendering. On this basis, we use the Unity 3D software for scene construction and management, as demonstrated in Figure~\ref{fig:8}.

We note again that the proposed DT framework comprises two main components: the data-driven optimization service on the cloud and the visualization application on the client side. During DT operation, the geometric model is loaded onto the client application once and for all. To reflect machine dynamics, the online data collected by the IoT gateway can be divided into two parts. The first part includes the operating parameters and monitoring data, which are directly transmitted to the Unity3D-based client application through the Socket tunnel, to meet the timeliness requirement of real-time rendering\footnote{The data tunnel between the gateway/cloud and the client application is implemented using C\# scripts.}. The rest of the data, such as process parameters and quality indicators, are first stored in the cloud database and then queried by the NN-based prediction module (see also Figure~\ref{fig:1}). Subsequently, the output of the optimization module is sent to the client application for updating the panel display.

\subsection{Generalizability of the DT Framework}
For the proposed DT framework, the visualization module requires field knowledge to construct the geometric model of different process production lines. Nevertheless, the other core modules, namely data processing and prediction/optimization service, can be easily adapted to  meet the production line control requirements in numerous scenarios. Specifically, the data processing module can be configured online to accommodate data streams of various protocols with different network sizes. The proposed composite NN model can be adapted (by altering the dimensions of the input/output layers) to encode a wide range of functional relationships between the inputs and outputs, as long as they are organized in time sequences. Meanwhile, with the help of parameter identification using sensitivity analysis, the AOA-based parameter search algorithm with a constrained search range can serve as a versatile tool for data-driven optimization, provided that the accuracy of the NN-based prediction model is ensured. The core advantage of our proposed DT framework lies in its data-driven nature, which enables it to adapt to diverse industry needs and operating conditions by re-identifying the input and output data, as well as the dimensions of the associated prediction and optimization models. 

\section{Results and Discussion}
\label{sec:Results}
\subsection{Experiment Parameters}
We evaluate the performance of our proposed DT framework in the scenario of ``thin-plate drying'' process on the tobacco shredding line, which has a total of 24 measurements and 1 timestamp. The measurements contain 22 parameters of the production process, including the moisture and temperature of the inlet material, exhaust air volume, roller wall temperature, exhaust damper opening level, hot air temperature, etc, which are collected from sensors deployed on different machines. There are 2 quality indicators, the moisture rate at the thin plate drying outlet and the processing strength of thin plate drying. The timestamp is transformed into four temporal codes in the input representation layer of the NN (see Figure~\ref{fig:4}).

For the proposed deep NN model, we set the ratio of the training set, validation set, and test set as 6:2:2, with 28 selected features and a time window of 32 samples as one single input.
The output channels of the 1D convolutional layer is set as 512. We adopt a TCN of 5 layers, with each layer having 512 channels for both the input and output. The convolution kernel size is set to 2, and the dilation distances between each element within the convolution kernel are set to $(1, 2, 4, 8, 16, 32)$. We choose a dropout rate of 0.25. The sparse self-attention layer is set to have 8 heads. The linear connections in the 2 fully connected layers are chosen as $(512, 64, 2)$ and $(32, 8, 1)$, respectively. The default learning rate is set as 0.001, the learning decay rate is set as 0.99, the training batch size is 128 and the number of iterations is 50. We choose Adam as the optimizer.

\subsection{Ablation Experiment of the Proposed NN Model}
This experiment aims to verify the effectiveness of the proposed TCN model with sparse self-attention in improving the accuracy of the prediction. The models designed for comparison in the ablation experiments include the Time encoding combined with TCN (T-TCN), TCN combined with the Probabilistic sparse self-Attention mechanism (TCN-PA), and T-TCN combined with Probabilistic sparse self-Attention mechanism (T-TCN-PA). T-TCN-PA corresponds to the full model of our propose composite NN. Under the same experimental conditions, the aforementioned models were trained, and the experimental results are presented in Table~\ref{Table:2}.
\begin{table}[t]
  \centering
  \caption{ Comparison of prediction errors of different models}
\begin{tabular}{ccccc}
\hline
 Algorithm & Quality indicators & MSE & MAE & $R^2$ \\ \hline
T-TCN & \begin{tabular}{@{}c@{}}moisture rate \\ Processing strength\end{tabular}  & \begin{tabular}{@{}c@{}}0.000601 \\ 0.001974\end{tabular} &  \begin{tabular}{@{}c@{}}0.021443 \\ 0.034870\end{tabular} & \begin{tabular}{@{}c@{}}0.935 \\ 0.921 \end{tabular} \\ \hline
TCN-PA &  \begin{tabular}{@{}c@{}}moisture rate \\ Processing strength\end{tabular}  & \begin{tabular}{@{}c@{}}0.000336 \\ 0.001686\end{tabular} &  \begin{tabular}{@{}c@{}}0.015016 \\ 0.030501\end{tabular} & \begin{tabular}{@{}c@{}}0.964 \\ 0.933 \end{tabular} \\ \hline
T-TCN-PA & \begin{tabular}{@{}c@{}}moisture rate \\ Processing strength\end{tabular}  & \begin{tabular}{@{}c@{}}0.000136 \\ 0.000922\end{tabular} &  \begin{tabular}{@{}c@{}}0.008754\\ 0.011675\end{tabular} & \begin{tabular}{@{}c@{}}0.986 \\ 0.981 \end{tabular} \\
 \hline
\end{tabular}
\label{Table:2}
\end{table}

To evaluate the prediction performance of the algorithms, three evaluation metrics, i.e., MSE, MAE, and $R^2$ score, for the two quality indicators were selected. From the comparison results in Table~\ref{Table:2}, it can be observed that among the three methods, the proposed composite NN model achieves the best prediction performance. With T-TCN-PA, the $R^2$ scores for the two quality indicators are 0.986 and 0.981, respectively, indicating that the algorithm's fitting scores are higher than those of the two other models. The MSEs are 0.000136 and 0.000922, respectively, indicating that its prediction errors are smaller than those of other models. This proves the superiority of the prediction performance by our proposed model, hence the effectiveness of the corresponding network structure.
\begin{figure}[t]
\includegraphics[width=0.9\linewidth]{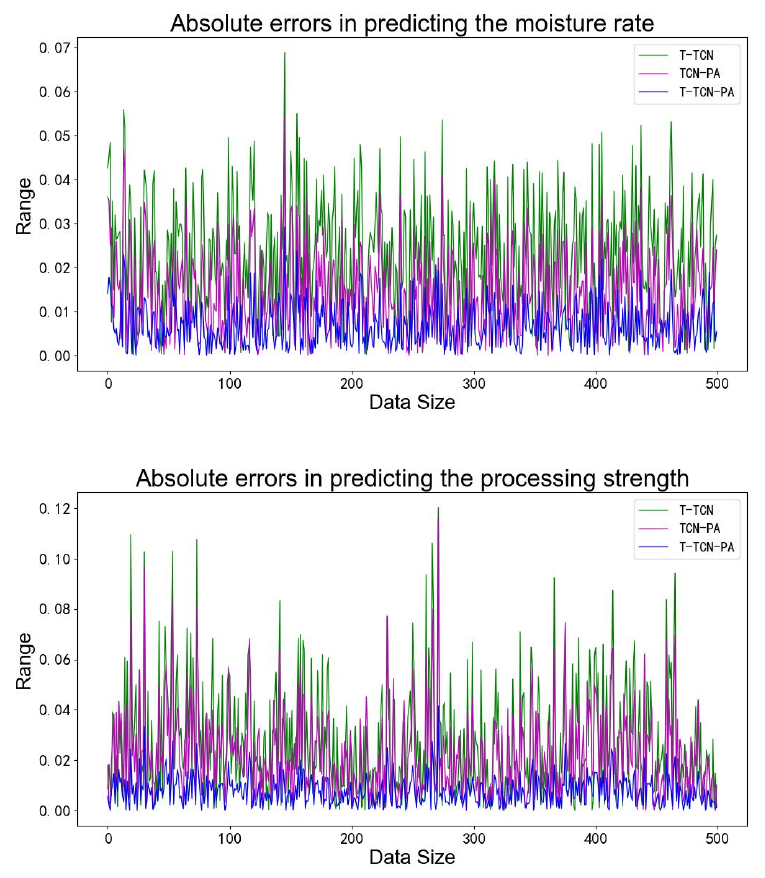}
\caption{\small{ Comparison of absolute error limits for prediction results with different network components.}}
 \label{fig:9}
\end{figure}

Figure~\ref{fig:9} compares the absolute prediction errors of three models with respect to the quality indicators of the moisture rate at the outlet and the processing strength. Observing Figure~\ref{fig:9}, it is clear that T-TCN-PA
0has the smallest error (the curve in blue) compared to the true quality values.

\subsection{Comparison Experiment between Different NN Models}
To further verify the efficiency of the proposed prediction model, comparative experiments are conducted using the canonical TCN and GRU networks, which have demonstrated good prediction performance in recent years. The same evaluation metrics, MSE, MAE and $R^2$, are used to assess the prediction performance of the algorithms. Under the same experimental conditions, the experimental results are presented in Table~\ref{Table:3}. Among the several methods compared, the proposed T-TCN-PA model exhibits the best prediction performance. T-TCN-PA achieves a higher accuracy for both quality indicators than the two reference models, demonstrating the effectiveness of the proposed network model.
\begin{table}[t]
  \centering
  \caption{ Comparison of prediction errors of different Networks}
\begin{tabular}{ccccc}
\hline
 Algorithm & Quality indicators & MSE & MAE & $R^2$ \\ \hline
TCN & \begin{tabular}{@{}c@{}}moisture rate \\ Processing strength\end{tabular}  & \begin{tabular}{@{}c@{}}0.009712 \\ 0.003017\end{tabular} &  \begin{tabular}{@{}c@{}}0.024154 \\ 0.043039\end{tabular} & \begin{tabular}{@{}c@{}}0.904 \\ 0.879 \end{tabular} \\ \hline
GRU &  \begin{tabular}{@{}c@{}}moisture rate \\ Processing strength\end{tabular}  & \begin{tabular}{@{}c@{}}0.000820 \\ 0.002261\end{tabular} &  \begin{tabular}{@{}c@{}}0.023198 \\ 0.037030\end{tabular} & \begin{tabular}{@{}c@{}}0.912 \\ 0.911 \end{tabular} \\ \hline
T-TCN-PA & \begin{tabular}{@{}c@{}}moisture rate \\ Processing strength\end{tabular}  & \begin{tabular}{@{}c@{}}0.000136 \\ 0.000922\end{tabular} &  \begin{tabular}{@{}c@{}}0.008754\\ 0.011675\end{tabular} & \begin{tabular}{@{}c@{}}0.986 \\ 0.981 \end{tabular} \\
 \hline
\end{tabular}
\label{Table:3}
\end{table}

\begin{figure}[t]
\includegraphics[width=0.9\linewidth]{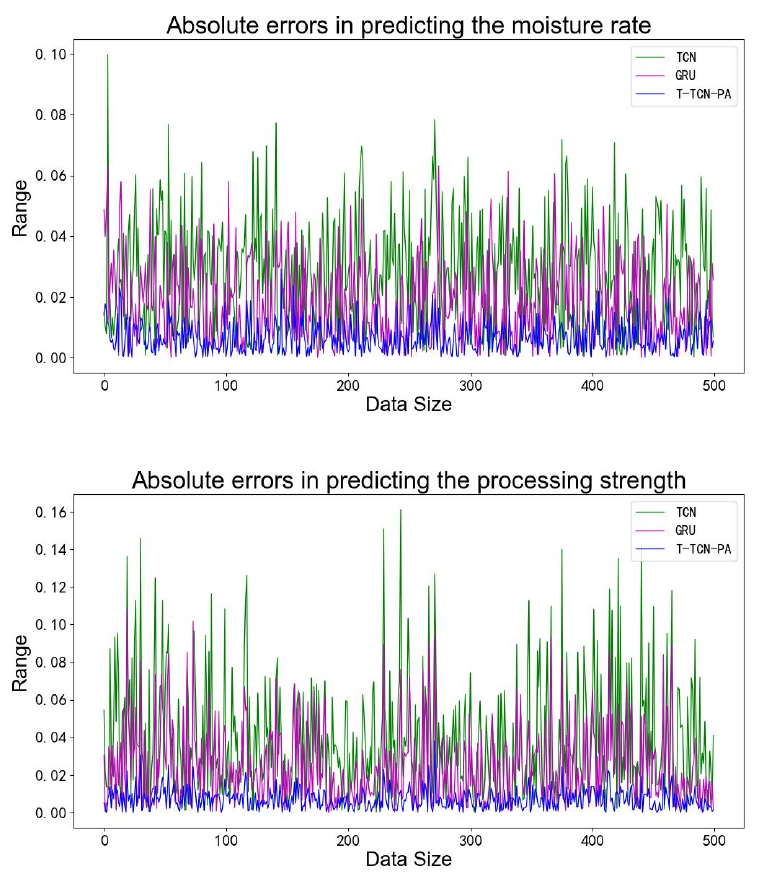}
\caption{\small{ Comparison of absolute error limits for prediction results with different network models.}}
 \label{fig:10}
\end{figure}

Figure~\ref{fig:10} compares the absolute prediction error of the three models in terms of the material moisture level at the outlet and the processing strength, respectively. It demonstrates that the proposed T-TCN-PA network has the smallest error (the curve in blue) against the true value of the quality indicators.

\subsection{Performance Analysis of the AOA-based Optimization Algorithm}

\begin{table}[t]
 \centering
 \caption{Categories of the 22 process parameters.}
\begin{tabular}{|p{2.1cm}|p{5.8cm}|}
\hline
Categories   & Parameters  \\ \hline
First category (non-controllable)  & Content moisture at inlet, inlet temperature,  ambient temperature, ambient humidity \\ \hline
Second category (non-independent) & Exhaust air volume, roller-wall temperature (front), roller-wall temperature (middle), roller-wall temperature (rear), main steam pressure, main steam temperature, hot air volume at the front chamber, average roller wall temperature, thin-plate steam temperature. outlet material temperature, negative pressure of the discharge hood  \\ \hline
Third category (controllable) & Exhaust damper opening level, hot air temperature, steam pressure after heating and humidifying, rear chamber damper opening level, thin plate steam pressure at inlet, front chamber damper opening level,  material flow rate at inlet  \\ \hline
\end{tabular}
\label{tab:4}
\end{table}

\begin{figure*}[t]
    \centering
    \begin{subfigure}{.24\textwidth}
        \centering
        \includegraphics[width=\linewidth]{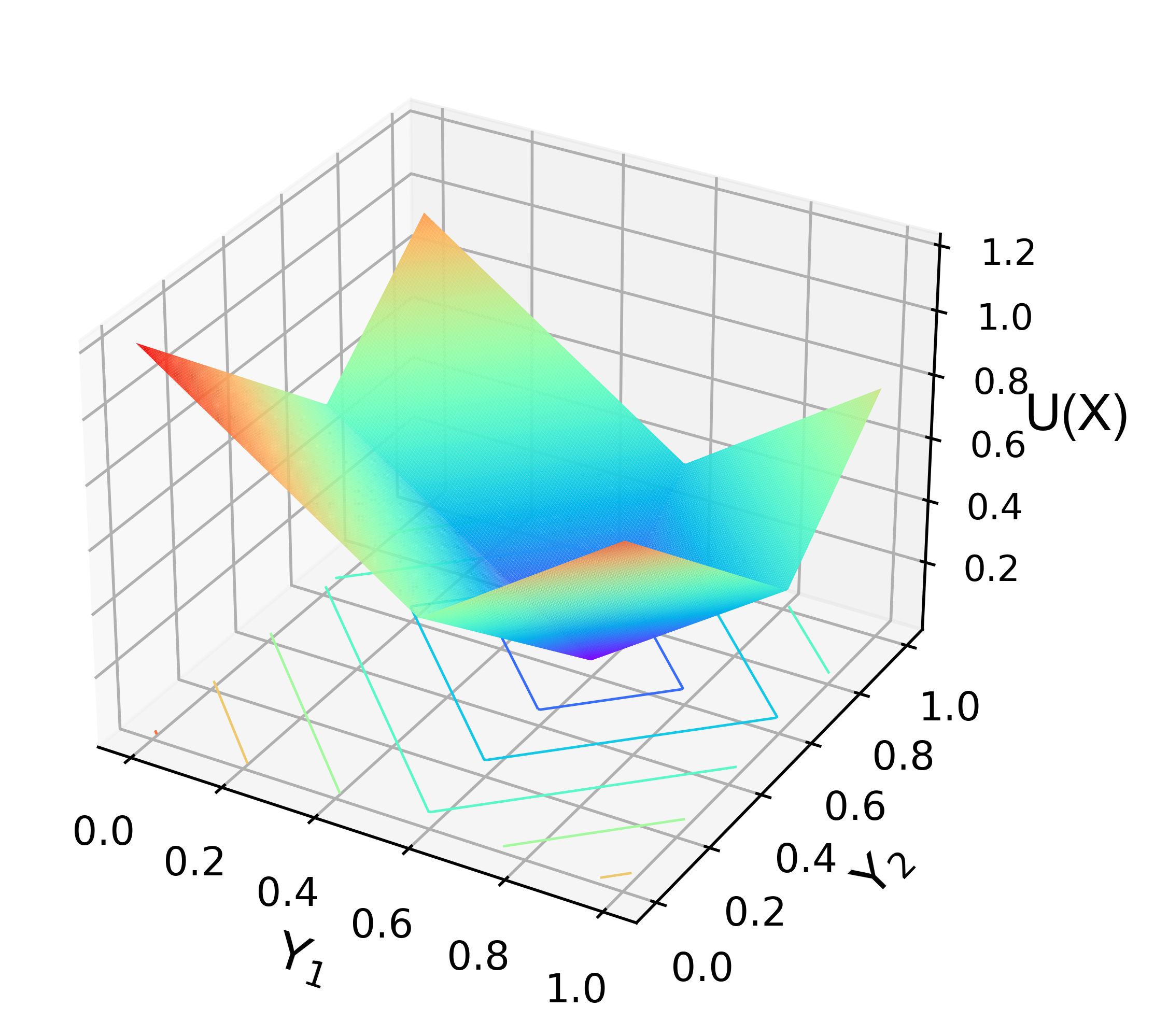}
                \caption{}
        \label{fig:11-1}
    \end{subfigure}%
    \begin{subfigure}{.24\textwidth}
        \centering
        \includegraphics[width=\linewidth]{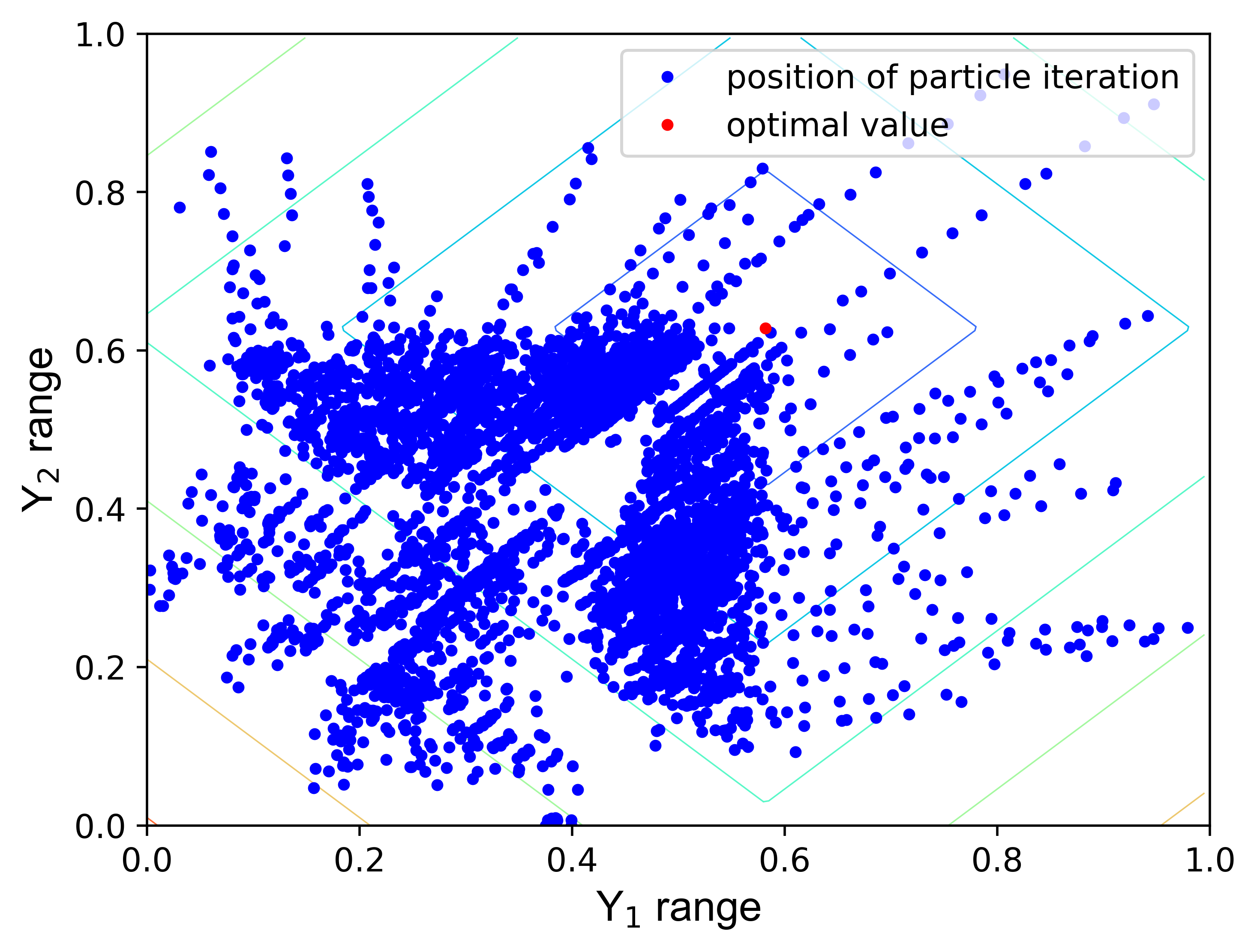}
                \caption{}
        \label{fig:11-2}
    \end{subfigure}%
    \begin{subfigure}{.24\textwidth}
        \centering
        \includegraphics[width=\linewidth]{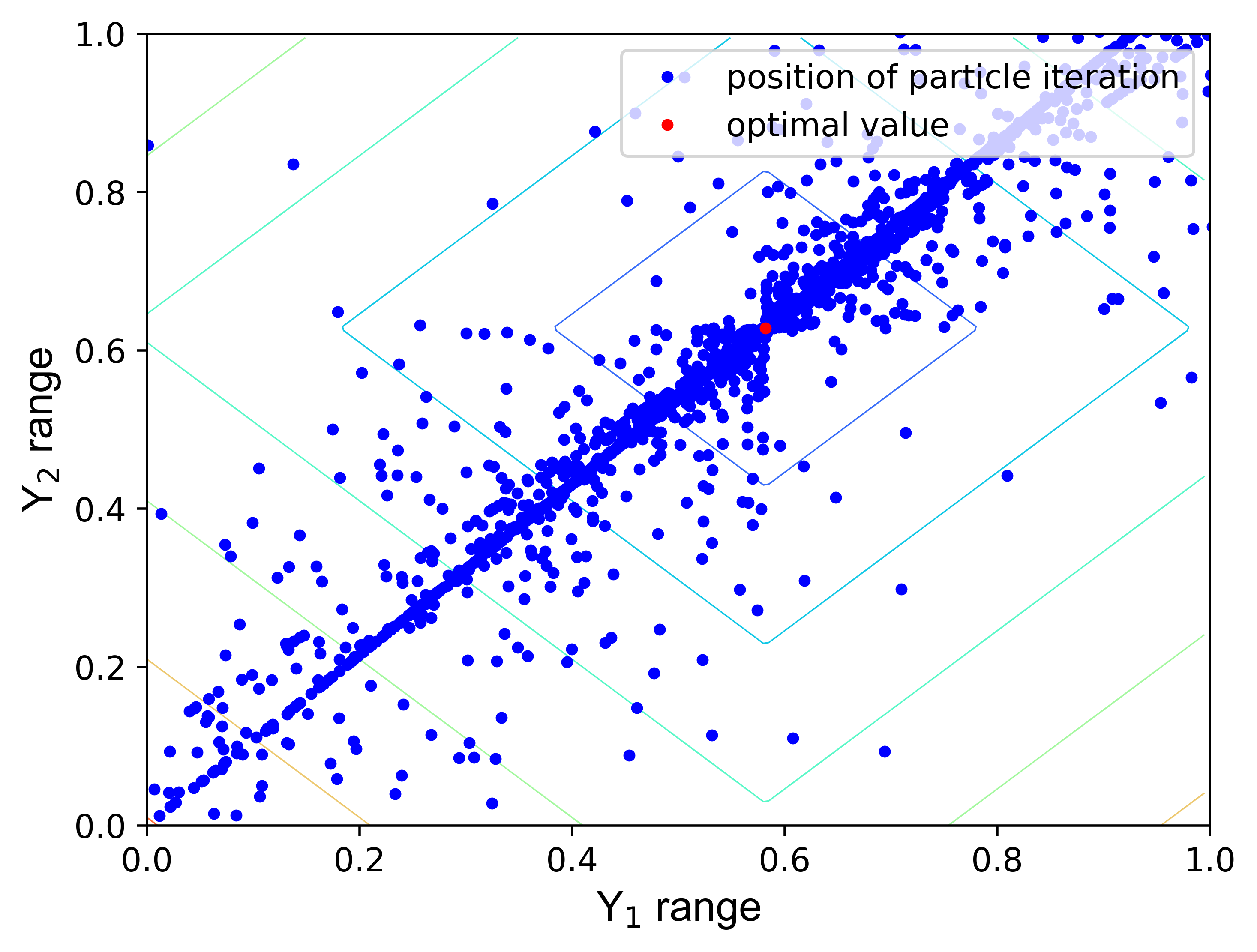}
                \caption{}
        \label{fig:11-3}
    \end{subfigure}%
    \begin{subfigure}{.24\textwidth}
        \centering
        \includegraphics[width=\linewidth]{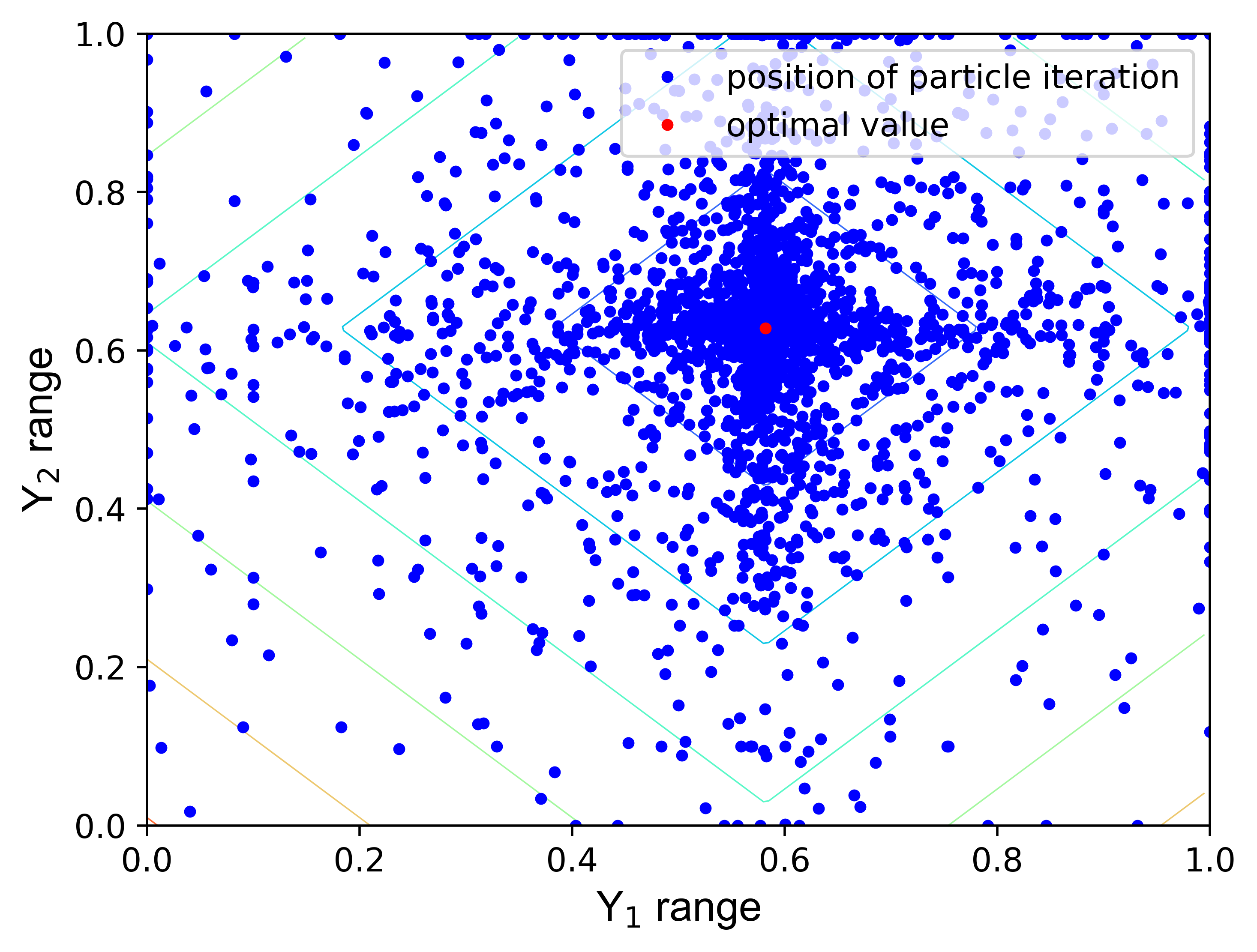}
                \caption{}
        \label{fig:11-4}
    \end{subfigure}

    \begin{subfigure}{.24\textwidth}
        \centering
        \includegraphics[width=\linewidth]{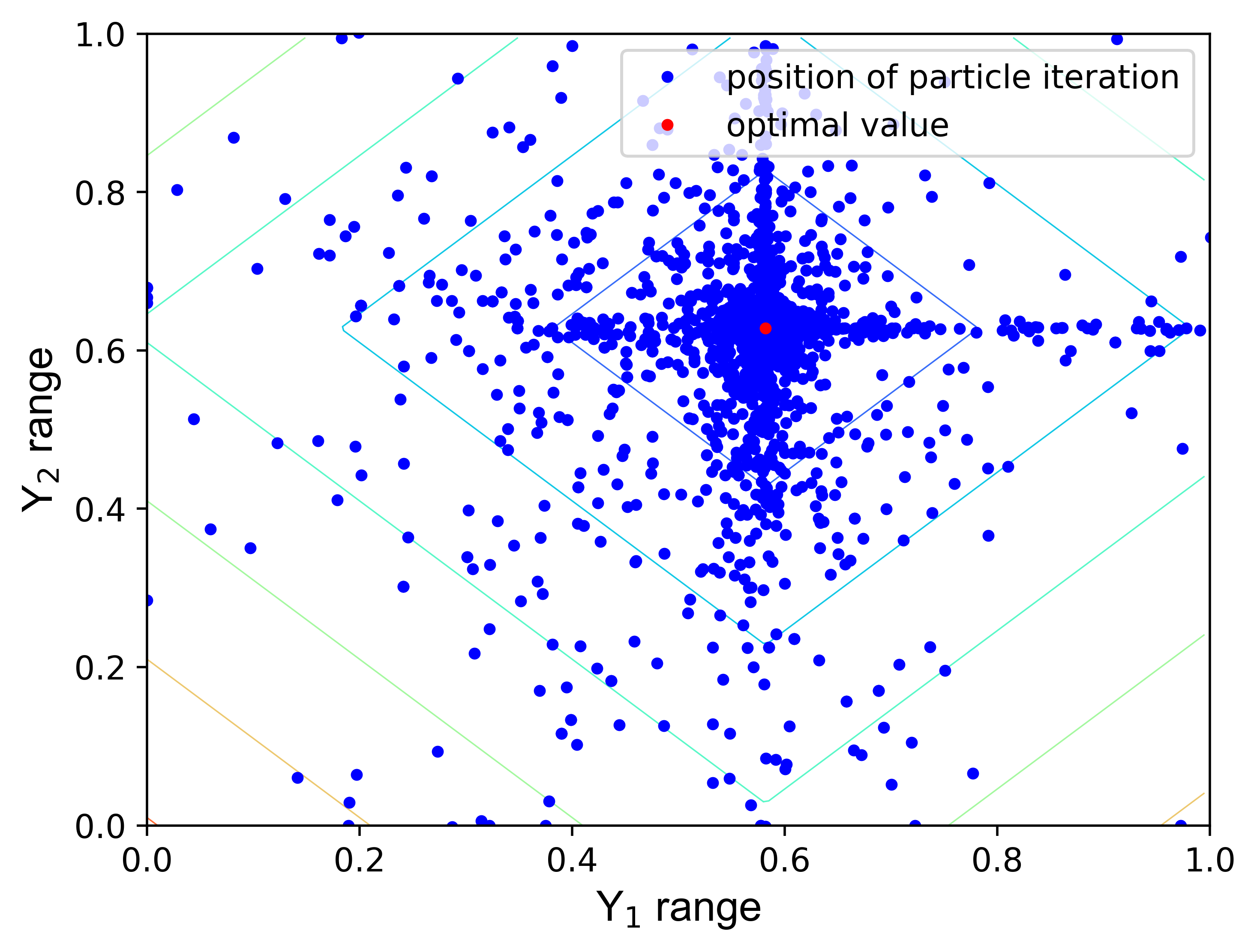}
                \caption{}
        \label{fig:11-5}
    \end{subfigure}%
    \begin{subfigure}{.24\textwidth}
        \centering
        \includegraphics[width=\linewidth]{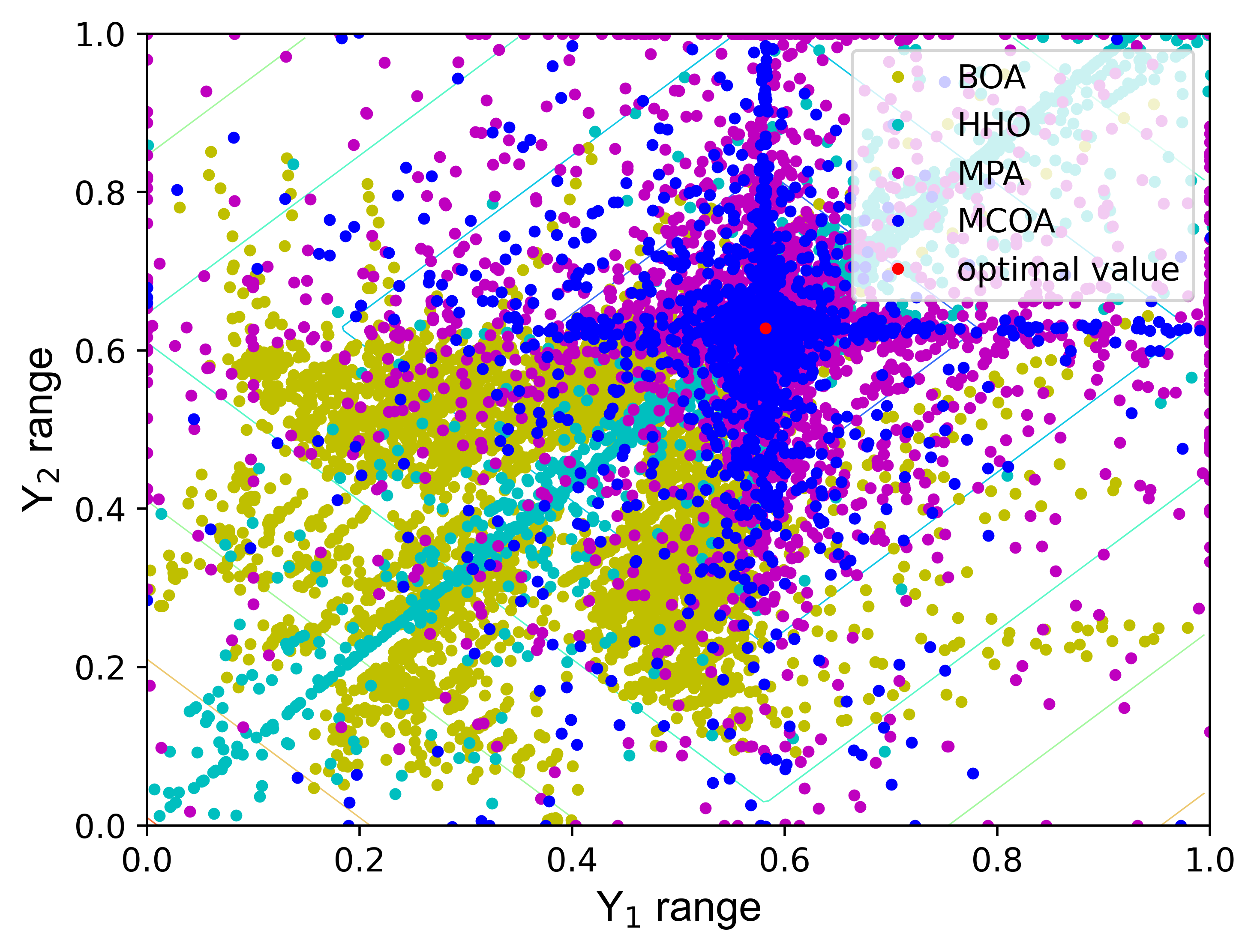}
        \caption{}
        \label{fig:11-6}
    \end{subfigure}%
    \begin{subfigure}{.24\textwidth}
        \centering
        \includegraphics[width=\linewidth]{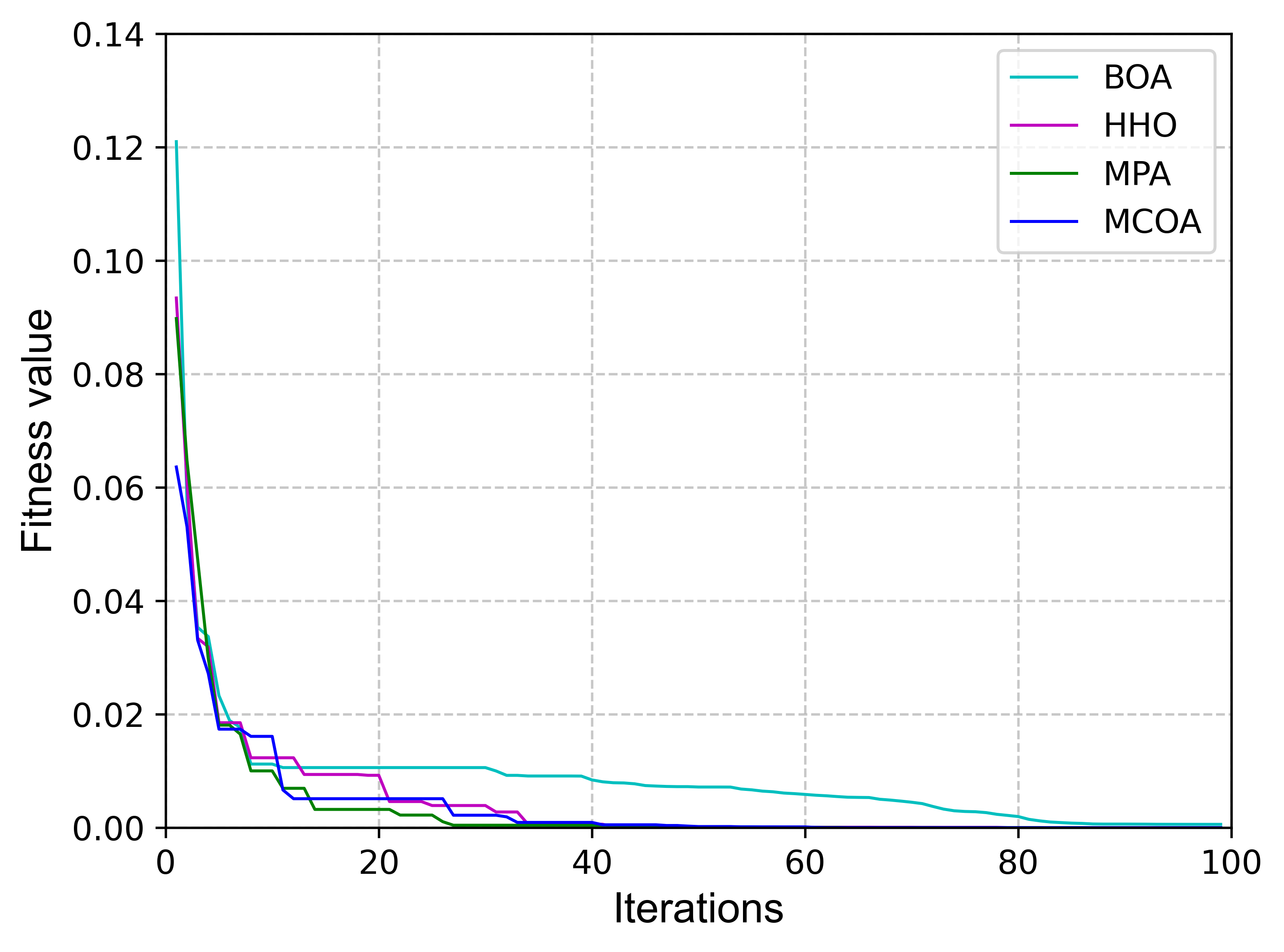}
                \caption{}
        \label{fig:11-7}
    \end{subfigure}%
    \begin{subfigure}{.24\textwidth}
        \centering
        \includegraphics[width=\linewidth]{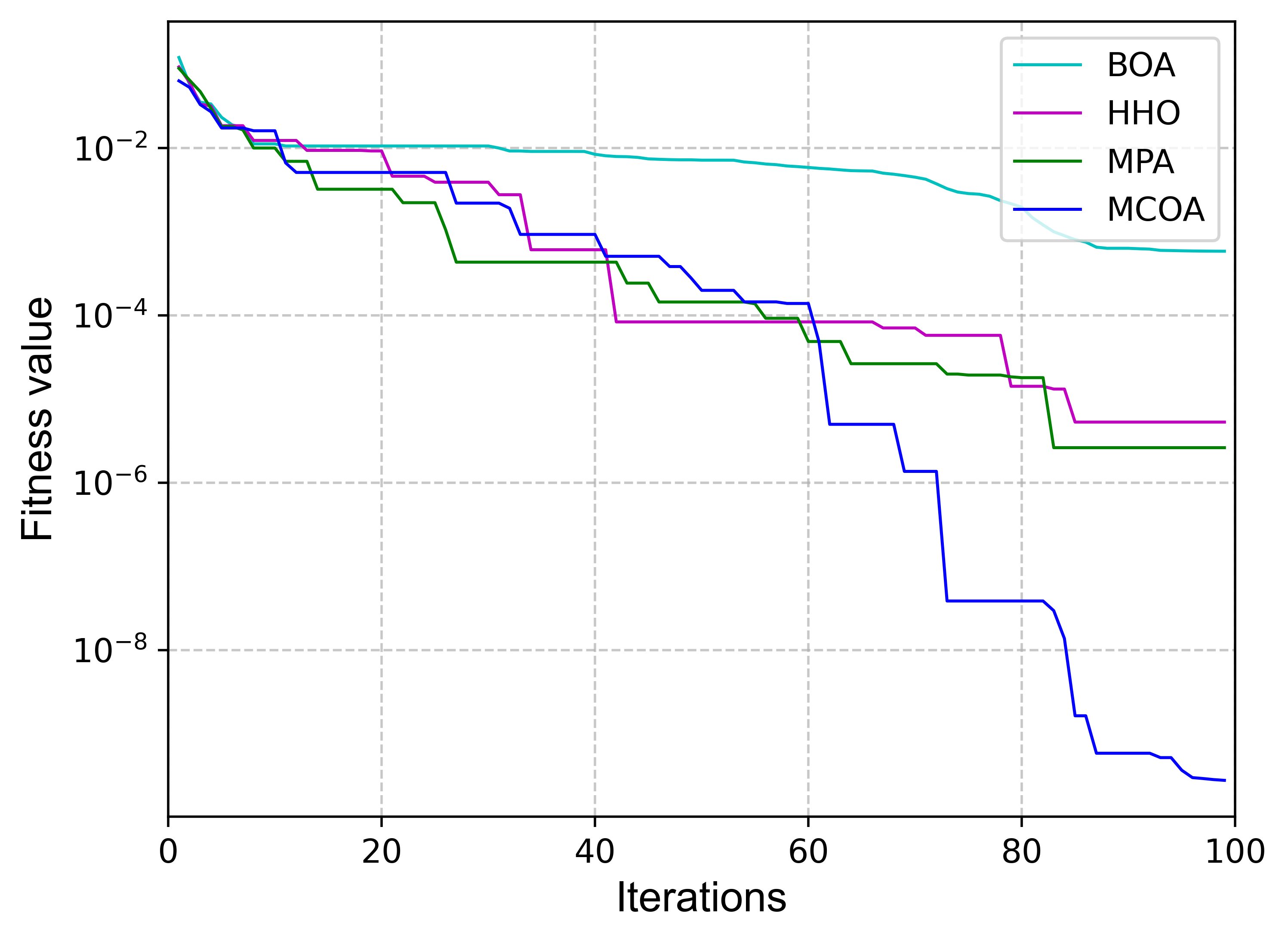}
        \caption{}
        \label{fig:11-8}
    \end{subfigure}
    \caption{\small{(a) Objective function. (b) Performance trajectory of BOA. (c) Performance trajectory of HHO. (d) Performance trajectory of MPA. (e) Performance trajectory of AOA. (f) Comparison of search trajectories. (g) Convergence over iterations for different algorithms. (h) Convergence comparison over iterations in logspace.}}
    \label{fig:11}
\end{figure*}

Applying sensitivity analysis in the actual production scenario (see also Section~\ref{sub_sec_extraction}), the process parameters are divided into three categories in Table~\ref{tab:4}. For the AOA algorithm deployed, a random number of particles is generated in different directions to search for the identified parameters of the third type. In the experiment, the performance of AOA is compared with that of the Butterfly Optimization Algorithm (BOA)~\cite{2019Butterfly}, the Harris Hawks optimization Algorithm (HHA)~\cite{HEIDARI2019849} and the Marine Predators Algorithm (MPA)~\cite{FARAMARZI2020113377}, which all demonstrate good performance in recent years. The target values for the two quality indicators are set as $(12.8\pm 0.2, 6.4\pm 0.1)$ according to the production requirements. For ease of comparison, the quality indicators are mapped to a normalized space, and the maximum iteration number is set to 100. All algorithms are run 25 times for comparison of average performance, whose results are shown in Table~\ref{tab:5} and Figure~\ref{fig:11}.
\begin{table}[t]
\centering
\caption{Performance comparison between 4 algorithms.}
\begin{tabular}{|p{2.2cm}|p{1.1cm}|p{1.1cm}|p{1.1cm}|p{1.1cm}|}
\hline
 & BOA & HHO & MPA & \bf{AOA} \\ \hline
Objective value (fitness) of the best solution in (\ref{eq_opt}) & $4.42 \times 10^{-6}$ & $7.22\times 10^{-13}$ &  $5.13\times 10^{-12}$ & $1.87\times 10^{-17}$ \\ \hline
Average objective value & $3.02\times 10^{-1}$ & $4.97\times 10^{-7}$ & $6.25\times 10^{-7}$ & $7.82\times 10^{-10}$ \\ \hline
Standard deviation & $1.94\times 10^{-1}$ & $2.85\times 10^{-7}$  & $2.54\times 10^{-7}$ & $3.65\times 10^{-10}$ \\ \hline
Success rate of adaptation & 52\% & 96\% & 92\% & 100\%\\ \hline
\end{tabular}
\label{tab:5}
\end{table}

Figure~\ref{fig:11-1} shows the value of the objective function (i.e., fitness, see (\ref{eq_opt})) obtained from exhaustive search in the normalized parameter space. Figures~\ref{fig:11-2}-\ref{fig:11-5} show the search trajectory (i.e., historical values of the two quality indicators) of the particles over 100 iterations for BOA, HHO, MPA, and the adopted AOA algorithm, respectively. Figure~\ref{fig:11-6} compares the historical particle optimization results of  the four algorithms over 100 iterations. Figure~\ref{fig:11-7} shows the convergence tendency of the four algorithms in terms of fitness as the number of iterations increases. Figure~\ref{fig:11-8} is a logspace representation of Figure~\ref{fig:11-7}. By observing Figures~\ref{fig:11-1}-\ref{fig:11-6}, it can be seen that the search path of the BOA algorithm diverges, making it difficult to find the optimal solution. The search path of HHO is diagonal and tends to overlook the optimal solution. Both MPA and the proposed AOA algorithm have search paths that contract from the periphery towards the center, but our proposed algorithm has a more concentrated convergence range and a faster convergence rate. From Figures~\ref{fig:11-7} and~\ref{fig:11-8}, it can be seen that the BOA algorithm struggles to find the optimal value for the objective function in concern, while both HHO and MPA can locate the optimal value. However, compared to the reference algorithms, the AOA algorithm achieves a higher fitness and a faster convergence speed.

Furthermore, we note that the accuracy of the NN-based prediction model has a direct influence on the optimization results of the process parameters, as it is utilized by the searching algorithm for objective function evaluation. To investigate this impact, we used different NN models that appeared in our previous experiments to simulate the objective value generation process. In this experiment, the process parameters obtained through the AOA-based search algorithm are implemented on the actual production line, and the Quality Acceptance Rates (QAR) for the products are compared. The solutions derived from the DT side are classified into two categories according to whether the actual product quality meets the requirement. A total of 30 tests were performed for comparison, as detailed in Table~\ref{table:7}. The findings indicate a significant decline in the accuracy of optimization results when the prediction model's accuracy falls below 92\% on average, while our proposed optimization method achieves a QAR of over 96\%.

\begin{table*}[t]
    \centering
    \caption{Comparison of QARs using different NN models in the searching process.}
    \begin{tabular}{cccccccc}
    \hline
    \multirow{2}{*}{Algorithm} & \multirow{2}{*}{Quality indicators} & \multicolumn{2}{c}{Qualified} & \multicolumn{2}{c}{Unqualified}&  \multirow{2}{*}{$R^2$} &  \multirow{2}{*}{QAR}\\ 
    \cline{3-6}
     & & MAE & MSE & MAE& MSE& \\ \hline
    \multirow{2}{*}{TCN}& moisture rate & 0.100414 & 0.012883 & 0.345361& 0.134156& 0.904& 53.3\%\\
    \cline{2-8}&Processing strength& 0.041968 & 0.002518 & 0.173426 & 0.033826 & 0.879& 50\%\\ \hline  
    \multirow{2}{*}{GRU}& moisture rate & 0.106695 & 0.014988 & 0.351296& 0.131160& 0.912& 56.7\%\\
    \cline{2-8}&Processing strength& 0.044646 & 0.002928 & 0.167300 & 0.031505 & 0.911& 56.7\%\\ \hline   
    \multirow{2}{*}{T-TCN}& moisture rate & 0.088957 & 0.010605 & 0.262815 & 0.073743 & 0.935& 83.3\%\\
    \cline{2-8}&Processing strength& 0.045358 & 0.002661 & 0.138493 & 0.019696 & 0.921& 80\%\\ \hline
    \multirow{2}{*}{TCN-PA}& moisture rate & 0.071358 & 0.006928 & 0.245951 & 0.063068 &0.964& 90\%\\
    \cline{2-8}&Processing strength& 0.043376 & 0.002647 & 0.136876 & 0.018881 &0.933& 86.7\%\\ \hline
    \multirow{2}{*}{T-TCN-PA}& moisture rate & 0.071152 & 0.007151 & / & / &0.986& 100\%\\
    \cline{2-8}&Processing strength& 0.038085 & 0.001981 & 0.130716 & 0.017086 &0.981& 96.7\%\\ \hline    
    \end{tabular}
    \label{table:7}
    \end{table*}    

\subsection{Stability Testing for the DT of Tobacco Shredding Line}
Because the visualization module is the sink of the data flow of the entire DT system, we can utilize the performance testing module provided by Unity3D for analyzing resource consumption (such as memory, CPU, GPU, and rendering rate) to measure the data-processing delay of the DT. After establishing a communication connection between the client application and the entire tobacco shredding line, 5 groups of comparative tests are conducted on the process of ``thin-plate drying''. Each group of tests records the average frame time used for the synchronization of the process data, the quality data, and the movement data of robotic arms on the client side. The stability of the DT system is evaluated through comparative analysis of the data in Table~\ref{tab:6}.
\begin{table}[t]
\centering
\caption{Data synchronization results for the DT. Frame interval time is recorded for synchronizing data of 3 types.}
\begin{tabular}{|p{0.8cm}|p{1.7cm}|p{2.1cm}|p{2.0cm}|}
\hline
Test Group & Synchronization time (ms) for process data &  Synchronization time (ms) for robotic movement data  & Synchronization time (ms) for quality prediction data \\ \hline
1 & 9.11 & 25.63 & 35.57 \\ \hline
2 & 8.98 &	26.22 &	35.66 \\ \hline
3 & 9.56 &	25.87	& 35.13 \\ \hline
4 & 9.42 &	25.99 &	35.94 \\ \hline
5 & 9.36 &	26.64 &	35.47 \\ \hline
\end{tabular}
\label{tab:6}
\end{table}

\begin{figure}[t]
	\centering
	\includegraphics[width=0.95\linewidth]{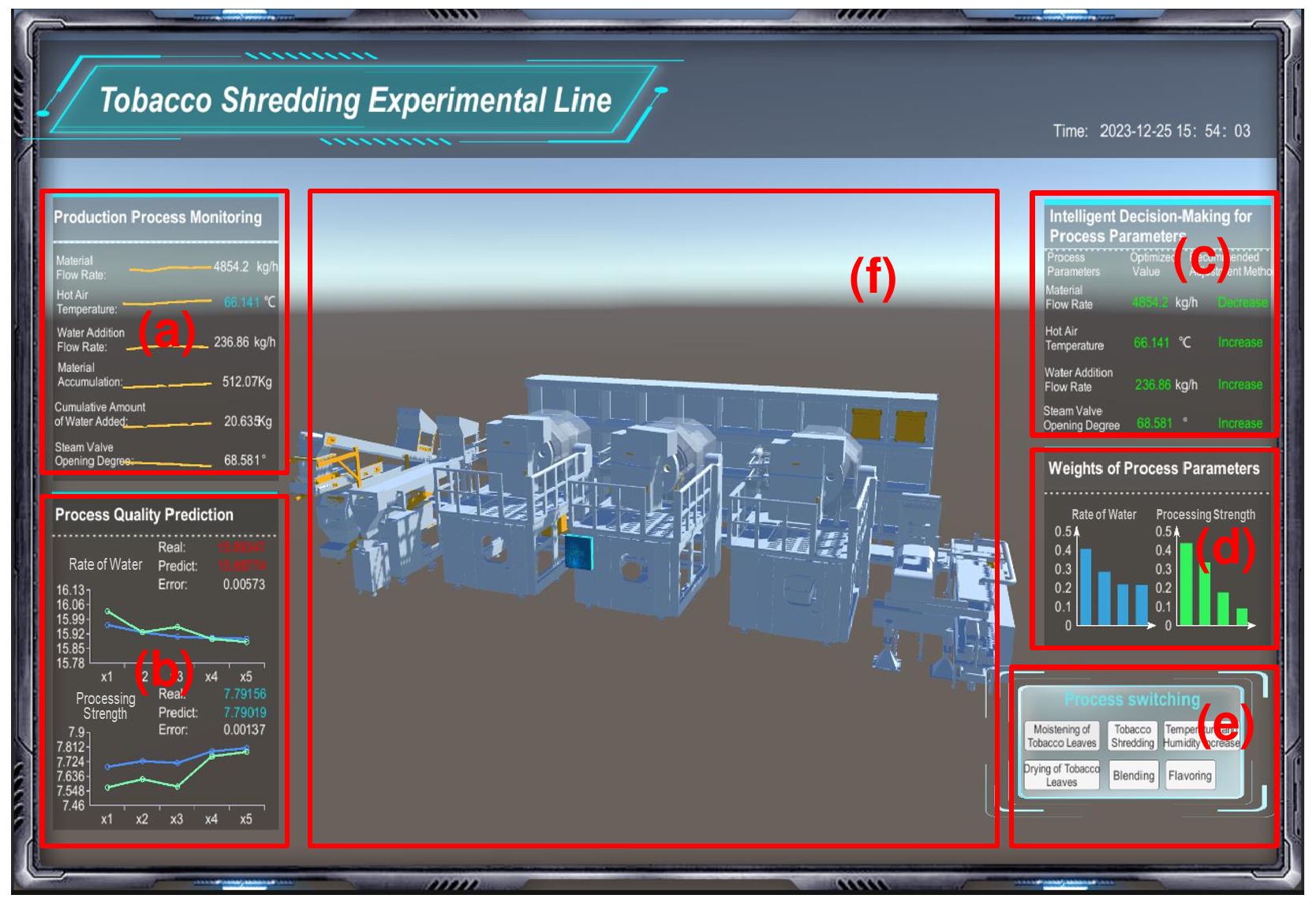}
	\caption{\small {The the DT production line on the client GUI. (a) Panel of production process states.  (b)  Panel of process quality predictions. (c)  Panel of recommended parameters after optimization. (d) Weights of the influence of process parameters on quality indicators. (e) Process switching board. (f) Real-time visualization of the production line.}}
	\label{fig:12}
\end{figure}

The average frame interval recorded in Table~\ref{tab:6} indicates the time taken for the visualization interface to refresh one frame with all data synchronized. Typically, human eyes need 25fps to perceive a smooth stream of images. By comparing the five sets of test data, it can be seen that the DT system works stably in terms of data synchronization. The relatively longer response time for synchronizing robot motion data is due to transmission of the relatively complex sensor/operation data from the production line to the client application. Prediction data take the longest time for synchronization, since the data is sent via backbone network from the cloud. It can be seen that negligible fluctuation occurs during data synchronization. The average time for all data synchronization translates to a frame rate of approximately 28fps, thus meeting the requirements for smooth visualization. The above test results demonstrate that the proposed DT system exhibits excellent visualization/motion response and low data delays. A snapshot of the data panel for the DT production line in the client application is shown in Figure~\ref{fig:12}.

Finally, it should be noted that the proposed DT framework eliminates most of the human factors from the control loop of the production line. Human feedback typically relies on engineers' observation and analysis of the production state, with response speed constrained by human reaction time and experience. In contrast, DT leverages real-time data analysis based on the proposed prediction and optimization models to swiftly detect and address abnormal situations. During field operation, the DT system demonstrates an average response time that is twice faster than humans, and the QAR of the product is reported to improve by 5\%.

\section{Conclusions}
\label{sec:Final}
To tackle the difficulty in providing an efficient control scheme of the process production lines, this paper propose a Digital Twin (DT)-based framework for product-quality prediction and real-time production parameter optimization. The DT provides a complete digital geometric mapping of the physical structure of the process production line. It also serves as a data-driven abstraction of the physical production line by mapping the functional relationship between parameters in the physical processes into the neural network-encoded input-output relationship in the virtual domain. Then, based on the real-time prediction of product quality using our proposed deep neural network, we have been able to provide advice on the optimal line parameter adjustment from the twin side to the physical side. Experiments demonstrate that our proposed system is able to achieve an average accuracy of 96\% for online product quality control.

The proposed DT-based quality control framework is easily adaptable to improve the production efficiency and process quality of various manufacturing lines. In future research, we plan to combine theories such as heat transfer and fluid mechanics with artificial intelligence methods to establish partially model-based digital twin production lines, aiming to further improve the accuracy and stability of DT under complex operating conditions.

\bibliographystyle{IEEEtran}
\bibliography{related}

\end{document}